\documentclass[runningheads]{llncs}
\usepackage{graphicx}
\usepackage{float}
\usepackage{tikz}
\usepackage{comment}
\usepackage{amsmath,amssymb} % define this before the line numbering.
\usepackage{color}
\usepackage{booktabs}
\usepackage{bbding}
\usepackage{subcaption}
\usepackage{makecell}
\usepackage{color}
\usepackage{multirow}
\definecolor{deepgreen}{rgb}{0,0.5,0}
\usepackage[pagebackref,breaklinks,colorlinks]{hyperref}
\usepackage{comment}
\usepackage{pifont}
\let\llncssubparagraph\subparagraph
\let\subparagraph\paragraph
\usepackage[small, compact]{titlesec}
\let\subparagraph\llncssubparagraph

\usepackage[capitalize]{cleveref}
\crefname{section}{Sec.}{Secs.}
\Crefname{section}{Section}{Sections}
\Crefname{table}{Table}{Tables}
\crefname{table}{Tab.}{Tabs.}

\def\eg{\emph{e.g.}}

\usepackage[accsupp]{axessibility}  % Improves PDF readability for those with disabilities.

\def\OURS{OOD-CV~}

\begin{document}
% \renewcommand\thelinenumber{\color[rgb]{0.2,0.5,0.8}\normalfont\sffamily\scriptsize\arabic{linenumber}\color[rgb]{0,0,0}}
% \renewcommand\makeLineNumber {\hss\thelinenumber\ \hspace{6mm} \rlap{\hskip\textwidth\ \hspace{6.5mm}\thelinenumber}}
% \linenumbers
\pagestyle{headings}
\mainmatter
\def\ECCVSubNumber{4514}  % Insert your submission number here

%\title{\OURS: A Benchmark for Robustness to Individual Nuisances in Real-World Out-of-Distribution Shifts} % Replace with your title
\title{\OURS: A Benchmark for Robustness to Out-of-Distribution Shifts of Individual Nuisances in Natural Images}

% INITIAL SUBMISSION 
%******************

% CAMERA READY SUBMISSION
%\begin{comment}
\titlerunning{A Benchmark for Robustness in Real-World OOD Shifts}
% If the paper title is too long for the running head, you can set
% an abbreviated paper title here
%
\author{Bingchen Zhao\inst{1}\and
Shaozuo Yu\inst{2}\and
Wufei Ma\inst{3}\and 
Mingxin Yu\inst{4}\and 
Shenxiao Mei\inst{3}\and
Angtian Wang\inst{3}\and 
Ju He\inst{3}\and
Alan Yuille\inst{3} \and 
Adam Kortylewski\inst{3,5,6}
%$^1$Tongji University\hspace{0.2em} $^2$ Johns Hopkins University\hspace{0.2em} $^3$ Purdue University\hspace{0.2em}  $^4$ Peking University\hspace{0.2em} $^5$ LunarAI\\
%{\tt\small \{zhaobc.gm, yu.shaozuo\}@gmail.com, \{ayuille1, akortyl1\}@jhu.edu}
}
%\author{First Author\inst{1}\orcidID{0000-1111-2222-3333} \and
%Second Author\inst{2,3}\orcidID{1111-2222-3333-4444} \and
%Third Author\inst{3}\orcidID{2222--3333-4444-5555}}
%
\authorrunning{B. Zhao et al.}
% First names are abbreviated in the running head.
% If there are more than two authors, 'et al.' is used.
%
\institute{$^1$University of Edinburgh
$^2$The Chinese University of Hong Kong\\
$^3$Johns Hopkins University 
$^4$Peking University\\
$^5$Max Planck Instutite for Informatics 
$^6$University of Freiburg
}
%Springer Heidelberg, Tiergartenstr. 17, 69121 Heidelberg, Germany
%\email{lncs@springer.com}\\
%\url{http://www.springer.com/gp/computer-science/lncs} \and
%ABC Institute, Rupert-Karls-University Heidelberg, Heidelberg, Germany\\
%\email{\{abc,lncs\}@uni-heidelberg.de}}
%\end{comment}
%******************
\maketitle

\begin{abstract}
Enhancing the robustness of vision algorithms in real-world scenarios is challenging. One reason is that existing robustness benchmarks are limited, as they either rely on synthetic data or ignore the effects of individual nuisance factors. We introduce \OURS, a benchmark dataset that includes out-of-distribution examples of 10 object categories in terms of pose, shape, texture, context and the weather conditions, and enables benchmarking models for image classification, object detection, and 3D pose estimation. 
In addition to this novel dataset, we contribute extensive experiments using popular baseline methods, which reveal that: 1) Some nuisance factors have a much stronger negative effect on the performance compared to others, also depending on the vision task. 2) Current approaches to enhance robustness have only marginal effects, and can even reduce robustness. 3) We do not observe significant differences between convolutional and transformer architectures. We believe our dataset provides a rich testbed to study robustness and will help push forward research in this area.
\end{abstract}

% TODO: semantic robustness?
% 
%%%%%%%%% BODY TEXT
\section{Introduction}
\label{sec:intro}

\begin{figure}
\centering
\includegraphics[width=\linewidth]{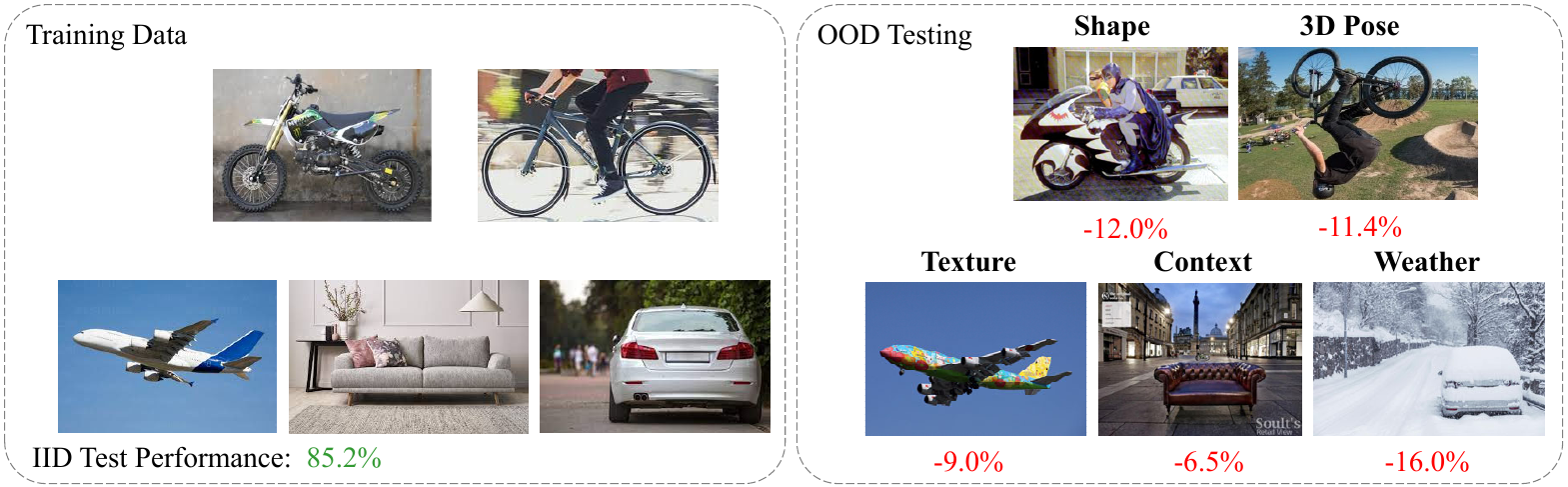}
\caption{Computer vision models are not robust to real-world distribution shifts at test time. For example, ResNet50 achieves $85.2\%$ accuracy on our benchmark, when tested on images that are similarly distributed as the training data (IID). However, its performance deteriorates significantly when individual nuisance factors in the test images break the IID assumption. Our benchmark makes it possible, for the first time, to study the robustness of image classification, object detection and 3D pose estimation to OOD shifts in individual nuisances.}%Our benchmark makes possible to study the robustness vision models to in.
%}
%, the classification performance drops significantly, we collected a set of images with annotated different nuisance factors and thus }
\label{fig:intro}
%\vspace{-0.7cm}
\end{figure}

Deep learning sparked a tremendous increase in the performance of computer vision systems over the past decade, under the implicit assumption that the training and test data are drawn independently and identically distributed (IID) from the same distribution. However, Deep Neural Networks (DNNs) are still far from reaching human-level performance at visual recognition tasks in real-world environments. The most important limitation of DNNs is that they fail to give reliable predictions in unseen or adverse viewing conditions, which would not fool a human observer, such as when objects have an unusual pose, texture, shape, or when objects occur in an unusual context or in challenging weather conditions (Figure \ref{fig:intro}). The lack of robustness of DNNs in such out-of-distribution (OOD) scenarios is generally acknowledged as one of the core open problems of deep learning, for example by the Turing award winners Yoshua Bengio, Geoffrey Hinton, and Yann LeCun \cite{bengio2021deep}. However, the problem largely remains unsolved.
% DL not robust 

One reason for the limited progress in OOD generalization of DNNs is the lack of benchmark datasets that are specifically designed to measure OOD robustness.
Historically, datasets have been pivotal for advancement of the computer vision field, e.g. in image classification \cite{deng2009imagenet}, segmentation \cite{lin2014microsoft,everingham2015pascal}, pose estimation \cite{xiang2014beyond,tremblay2018falling,xiang2017posecnn}, and part detection \cite{chen2014detect}.
%A recent approach to foster the development of vision systems that are robust was made by the organization of a Robust Vision Challenge \cite{rovi}. The organizers tested the robustness of computer vision algorithms by evaluating them across different datasets. While measuring the cross-dataset generalization of algorithms can be a proxy for model robustness, it does not allow for the quantification of robustness w.r.t. different nuisance variables.
However, benchmarks for OOD robustness have important limitations, which limit their usefulness for real-world scenarios. Limitations of OOD benchmarks can be categorized into three types:
Some works measure robustness by training models on one dataset and testing them on another dataset without fine-tuning \cite{rovi,ye2021ood,hendrycks2019robustness,hendrycks2021nae}. 
However, cross-dataset performance is only a very coarse measure of robustness, which ignores the effects of OOD changes to individual nuisance factors such as the object texture, shape or context.
Other approaches artificially generate corruptions of individual nuisance factors, such as weather \cite{michaelis2019dragon}, synthetic noise \cite{hendrycks2019robustness} or partial occlusion \cite{wang2020robust}.
However, some nuisance factors are difficult to simulate, such as changes in the object shape or 3D pose. Moreover, artificial corruptions only have limited generalization ability to real-world scenarios.
The third type of approach obtains detailed annotation of nuisance variables by recording objects in fully controlled environments, such as in a laboratory \cite{ilab} or using synthetic data \cite{kortylewski2018empirically}. But such controlled recording can only be done for limited amount of objects and it remains unclear if the conclusions made transfer to real-world scenarios. 

In this work, we introduce \OURS, a dataset for benchmarking OOD robustness on real images with annotations of individual nuisance variables and labels for several vision tasks. Specifically, the training and IID testing set in \OURS consists of $10$ rigid object categories from the PASCAL VOC 2012 \cite{pascal-voc-2012} and ImageNet \cite{deng2009imagenet} datasets, and the respective labels for image classification, object detection, as well as the 3D pose annotation from the PASCAL3D+ dataset \cite{xiang2014beyond}.
Our main contribution is the collection and annotation of a comprehensive out-of-distribution test set consisting of images that vary w.r.t. the training data in PASCAL3D+ in terms individual nuisance variables, i.e. images of objects with an unseen shape, texture, 3D pose, context or weather (\cref{fig:intro}). 
Importantly, we carefully select the data such that each of our OOD data samples only varies w.r.t. one nuisance variable, while the other variables are similar as observed in the training data. 
We annotate data with class labels, object bounding boxes and 3D object poses, resulting in a total dataset collection and annotation effort more than 650 hours.
Our ROBIN dataset, for the first time, enables studying the influence of individual nuisances on the OOD performance of vision models. 
In addition to the dataset, we contribute an extensive experimental evaluation of popular baseline methods for each vision task and make several interesting observations, most importantly:
1) Some nuisance factors have a much stronger negative effect on the model performance compared to others. Moreover, the negative effect of a nuisance depends on the downstream vision task, because different tasks rely on different visual cues.
2) Current approaches to enhance robustness using strong data augmentation have only marginal effects in real-world OOD scenarios, and sometimes even reduce the OOD performance. Instead, some results suggest that architectures with 3D object representations have an enhanced robustness to OOD shifts in the object shape and 3D pose.
3) We do not observe any significant differences between convolutional and transformer architectures in terms of OOD robustness.
%3) Deep networks are much more sensitive to changes in the object pose, shape and weather than to changes in context and the object texture. 
%4) Finer-grained vision tasks, such as 3D pose estimation, are more challenging in OOD scenarios compared to classification and detection.
%5) Adversarial training has no effect on real-world robustness.
We believe our dataset provides a rich testbed to benchmark and discuss novel approaches to OOD robustness in real-world scenarios and we expect the benchmark to play a pivotal role in driving the future of research on robust computer vision.
% Developments in cv have been driven by benchmark datasets, 
% but robustness Benchmarks are not good, e.g. OOD-bench\cite{ye2021ood}
% leading to some people claiming that the gap between humans and machines is closing, while in fact it is still very big
% We present new benchmark / introduce details here and show in Figure 1
% Our experiments show interesting things:
%

\section{Related works}
%\vspace{-0.5cm}
%\begin{table}
%    \centering
%    \setlength\tabcolsep{0.3em}
%    \begin{tabular}{lccc}
%    \toprule
%    Dataset  & Image Source  & Task Type & Individual Nuisances        \\ 
%    \midrule
%    ImageNet-C~\cite{hendrycks2019robustness} & Synthetic          &  Cls &  \textcolor{deepgreen}{\cmark}  \\ 
%    COCO-C~\cite{michaelis2019dragon}     &    Synthetic       &     Det &    \textcolor{deepgreen}{\cmark}     \\
%    ImageNet-A~\cite{hendrycks2021nae} & Real-World       & Cls       &    \textcolor{red}{\xmark}     \\
%    ImageNet-R~\cite{hendrycks2021many} &  Real-World      &  Cls    &    \textcolor{red}{\xmark}  \\
%    \midrule
%    \OURS      &  Real-World      &   Cls,Det,3dp & \textcolor{deepgreen}{\cmark}     \\ 
%    \bottomrule
%    \end{tabular}
%    \caption{\label{tab:dataset_comparison}Comparison of our dataset to existing robustness benchmarks. 
%    ''Cls``, ''Det``, and ''3dp`` stands for Image Classification, Object Detection, and 3D pose estimation respectively.}
%    \vspace{-1cm}
%\end{table}
% \cmark, \xmark
%\todo{OOD-BENCH\cite{ye2021ood}}

\noindent \textbf{Robustness benchmark on synthetic images.}~~
%Adversarial examples are the one of the most interesting properties of DNNs, and adversarial robustness consider the robustness of neural network faced with artificial adversarial examples which are generated by searching for the smallest addition perturbation in the RGB space that is enough to cause the neural network make wrong decisions~\cite{kurakin2016adversarial}.
There has been a lot of recent work on utilizing synthetic images to test the robustness of neural networks~\cite{kurakin2016adversarial,hendrycks2019robustness,michaelis2019dragon}.
For example, ImageNet-C~\cite{hendrycks2019robustness} evaluates the performance of neural networks on images with synthetic noises such as JPEG compression, motion-blur and Gaussian noise by perturbing the standard ImageNet~\cite{deng2009imagenet} test set with these noises. 
\cite{michaelis2019dragon} extends this idea of perturbing images with synthetic noises to the task of object detection by adding these noises on COCO~\cite{lin2014microsoft} and Pascal-VOC~\cite{everingham2015pascal} test sets.
Besides perturbation from image processing pipelines, there are also work~\cite{geirhos2018} benchmarks the shape and texture bias of DNNs using images with artificially overwritten textures.
Using style-transfer~\cite{gatys2016image} as augmentation~\cite{geirhos2018} or using a linear combination between strongly augmented images and the original images~\cite{hendrycks2019augmix} have been shown as effective ways of improving the robustness against these synthetic image noises or texture changes.
However, these benchmarks are limited in a way that synthetic image perturbations are not able to mimic real-world 3-dimensional nuisances such as novel shape or novel pose of objects.
Our experiments in~\cref{sec:exp} also show that style-transfer~\cite{gatys2016image} and strong augmentation~\cite{hendrycks2019augmix} does not help with shape and pose changes.
In addition, these benchmarks are limited to single tasks, for example, ImageNet-C~\cite{hendrycks2019robustness} only evaluates the robustness on image classification, COCO-C~\cite{michaelis2019dragon} only evaluates on the tasks of object detection.
DomainBed~\cite{gulrajani2020search} also benchmarks algorithm on OOD domain generalization on the task of classification.
In our work, we evaluate the robustness on real world images, while also evaluate the robustness across different tasks including image classification, object detection, and pose estimation.

\noindent \textbf{Robustness benchmark on real world images.}~~
%Benchmark the robustness of neural network on a test set of real world images has more practical values for deploying computer vision models in the real world. 
%After all, those real world image are naturally occurring examples that make the model performance to degrade, and is more likely to be seen in real scenarios.
Distribution shift in real-world images are more than just synthetic noises, many recent works~\cite{recht2019imagenet,hendrycks2021nae,hendrycks2019robustness} focus on collecting real-world images to benchmark robustness of DNN performances.
ImageNet-V2~\cite{recht2019imagenet} created a new test set for ImageNet~\cite{deng2009imagenet} by downloading images from Flickr, and found this new test set causes the model performance to degrade, showing that the distribution shift in the real images has an important influence on DNN models.
%ImageNet-V2~\cite{recht2019imagenet} first shows that the performance of state-of-the-art models degrade to real images that has distribution shifts.
By leveraging an adversarial filtration technique that filtered out all images that a fixed ResNet-50~\cite{he2015deep} model can correctly classifies, ImageNet-A~\cite{hendrycks2021nae} collected a new test set and shows that these adversarially filtered images can transfer across other architectures and cause the performance to drop by a large margin.
Although ImageNet-A~\cite{hendrycks2021nae} shows the importance of evaluating the robustness on real-world images, but cannot isolate the nuisance factor. %and thus cannot give us many insights on how the neural network fails and how to improve current vision models.
Most recently, ImageNet-R~\cite{hendrycks2021many} collected four OOD testing benchmarks by collecting images with distribution shifts in texture, geo-location, camera parameters, and blur respectively, and shows that not one single technique can improve the model performance across all the nuisance factors.
There are also benchmarks to test how well a model can learn invariant features from unbalanced datasets~\cite{tang2022invariant}.
And benchmarks composed of many real world shifts~\cite{koh2021wilds}.
We introduce a robustness benchmark that is complementary to prior datasets, by disentangling individual OOD nuisance factors that correspond to semantic aspects of an image, such as the object texture and shape, the context object, and the weather conditions. Due to rich annotation of our data, our benchmark also enables studying OOD robustness for various vision tasks.

\noindent \textbf{Techniques for improving robustness.}~~
To close the gap between the performance of vision models on datasets and the performance in the real-world, many techniques has been proposed~\cite{Mohseni2021PracticalML}. 
These techniques for improving robustness can be roughly categorized into two types: data augmentation and architectural changes.
Adversarial training by adding the worst case perturbation to images at training-time~\cite{wong2020fast}, using stronger data augmentation~\cite{cubuk2018autoaugment,wang2021augmax}, image mixtures~\cite{hendrycks2019augmix,yun2019cutmix,erichson2022noisymix}, and image stylizations~\cite{geirhos2018} during training, or augmenting in the feature space~\cite{hendrycks2021many} are all possible methods for data augmentation.
These data augmentation methods have been proven to be effective for synthetic perturbed images~\cite{hendrycks2019augmix,geirhos2018}.
Architectural changes are another way to improve the robustness by adding additional inductive biases into the model.
\cite{xie2019feature} proposed to perform de-noise to the feature representation for a better adversarial robustness. 
Analysis-by-synthesis appoaches~\cite{wang2021nemo,kortylewski2021compositional} can handle scenarios like occlusion by leveraging a generative object model and through top-down feedback \cite{xiao2020tdmpnet}.
Transformers are a newly emerged architecture for computer vision~\cite{dosovitskiy2020image,liu2021swin,Shao_2021_WACV}, and there are works showing that transformers may have a better robustness than CNNs~\cite{bhojanapalli2021understanding,mahmood2021robustness}, although our experiments suggest that this is not the case.
Object-centric representations~\cite{locatello2020object,wen2022selfsupervised} have also been show to improve robustness.
Self-supervised learned representations also show improvement on OOD examples~\cite{hendrycks2019selfsupervised,zhao2020distilling,zhu2021improving,cui2022discriminabilitytransferability}
Our benchmark enables the comprehensive evaluation of such techniques to improve the robustness of vision models on realistic data, w.r.t. individual nuisances and vision tasks. We find that current approaches to enhance robustness have only marginal effects, and can even reduce robustness, thus highlighting the need for an enhanced effort in this research direction. 
%Although progress has been made for improving the robustness of vision models, most of the techniques can only improve one aspect of the model's robustness, in our benchmark, we aim to provide a more comprehensive evaluation of semantic robustness and thus promote the research of technique for improve the robustness.

\begin{figure*}
    \centering
    \includegraphics[width=0.95\textwidth]{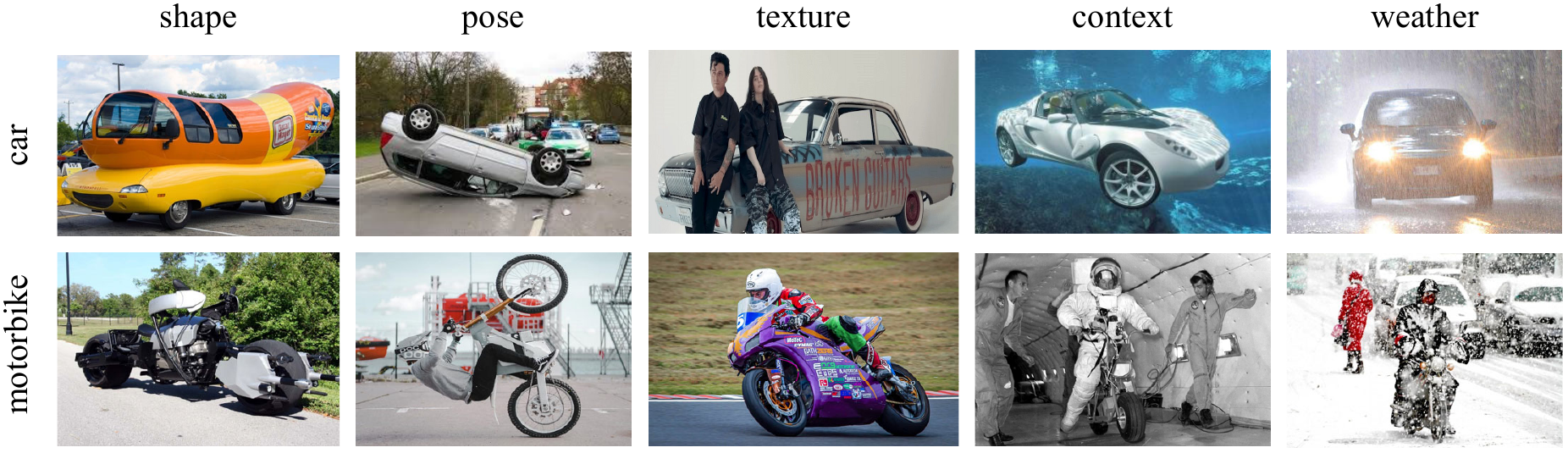}
    \caption{Examples from our dataset with OOD variations of individual nuisance factors including the object shape, pose, texture, context and weather conditions.
    %Modified batblade and special vehicles have an unusual shape compared to any ordinary motorbike or cars in the training set. At accident scenes or extreme sports peculiar pose can appear, and novel textures can happen on modified vehicles. Objects can also appear in strange context such as in the water or space labs. Weather conditions are another factor that can influence all the pixels in an image and causing the performance of the model to degrade.
    }
    \label{fig:data_demo}
%\vspace{-.7cm}
\end{figure*}

\section{Dataset collection}
\label{sec:collection}
In this section, we introduce the design of the \OURS benchmark and discuss the data collection process to obtain the OOD images and annotations.

\subsection{What are important nuisance factors?}
\label{sec:ontology}
The goal of the \OURS benchmark is to measure the robustness of vision models to realistic OOD shifts w.r.t. important individual nuisance factors.
To achieve this, we define an ontology of nuisance factors that are relevant in real-world scenarios following related work on robust vision \cite{michaelis2019benchmarking,rosenfeld2018elephant,qiu2016unrealcv,alcorn2019strike,kortylewski2019analyzing} and taking inspiration from the fact that images are 3D scenes with a hierarchical compositional structure, where each component can vary independently of the other components. 
%, of which some also been studied in related 
%, we take a compositional 3D scene representation into account~\cite{niemeyer2021giraffe}, \ie, 
In particular, we identify five important nuisance factors that vary strongly in real-world scenarios: the object shape, its 3D pose, and texture appearance, as well as the surrounding context and the weather conditions.
These nuisance factors can be annotated by a human observer with reasonable effort, while capturing a large amount of the variability in real-world images.
Notably, each nuisance can vary independently from the other nuisance factors, which will enable us to benchmark the OOD effect of each nuisance individually. 
%images are composed of 3D objects and the surrounding context, and 
%So we define these four attributes of a scene, namely shape, pose, texture, and context as the variable we want to control.
%Besides this four attributes, we also identify another variable that can occur in the real-world and have a great influence of the appearance of the image - weather, which can be treated as a variable independent of the previous four attributes.
%In summary, we aim to collect images with the above five individual nuisance factors, \ie, images with novel object shape, pose, or texture, and image with novel context or weather.

\subsection{Collecting images}
%\vspace{-0.1cm}
%Specifically, we want each of images collected has only one variable that is out-of-distribution while the other variable are the same with the training distribution.
OOD data can only be defined w.r.t. some reference distribution of training data.  
For our dataset, the reference training data is based on the PASCAL3D+~\cite{xiang_wacv14} dataset which is composed of images from Pascal-VOC~\cite{everingham2015pascal} and ImageNet~\cite{deng2009imagenet} datasets, and contains annotations of the object class, bounding box and 3D pose. 
Our goal is to collect images where only one nuisance factor is OOD w.r.t. training data, while other factors are similar as in training data.

To collect data with OOD nuisance factors, we search the internet using a curated set of search keywords that are combinations of the object class from the PASCAL3D+ dataset and attribute words that may retrieve images with OOD attributes, e.g. "car+hotdog" or "motorbike+batman", a comprehensive list of our search keywords used can be found in the supplementary material.
Note that we only use $10$ object categories from PASCAL3D+ %(Table \ref{tab:stats_per_cls})
, as we could not find sufficient OOD test samples for all nuisances for the categories "bottle" and "television".
%\todo{put the class names in supp}
%making the objects rarely seen, such as `batmobile' for pictures of cars modified to mimic the car of Batman from the Batman movies and comices.
%There may be false positives with no novel attributes and also images containing multiple nuisance factors. To ensure that each image in our benchmark only has one OOD variable, 
We manually filtered images with multiple nuisances and put an effort in retaining images that significantly vary in terms of one nuisance only.
% \todo{put the images with no nuisance or multiple nuisances in the supp}
Following this approach, we collect $2632$ images with OOD nuisances in terms of shape, texture, context and weather. 
Examples images are shown in~\cref{fig:data_demo}.

%Using the 3D pose and shape annotation from PASCAL3D+, we can create OOD splits of these nuisance variables without collecting additional data.
To create OOD dataset splits regarding 3D pose and shape we leverage the shape and pose annotations from PASCAL3D+. These allow us to split the dataset such that 3D pose and shape of training and testing set do not overlap. 
We augment these OOD splits in pose and shape with additional data that we collect from internet. Statistics of our dataset are shown in the supplementary.%Table \ref{tab:stats_per_cls}. 
On average we have $52$ images per nuisance and object class which is comparable to other datasets, e.g. ImageNet-C with an average of $50$ images.

Overall, the \OURS benchmark is an image collection with a total of $13297$ images composed from PASCAL3D+ and internet where $10665$ images are from PASCAL3D+ and $2632$ images are collected and annotated by us. To ensure that test data is really OOD, three annotators went through all training data from PASCAL3D+ and filtered out images from training set that were too similar to OOD test data. 
%We also screened the test data before the annotation to ensure that no additional biases are introduced, e.g. due to two sources of images.
To enable us to benchmark OOD robustness, the nuisance factors and vision tasks were annotated as discussed in the next section.

\begin{comment}

\begin{table}
%\vspace{-.7cm}
    \centering
    \caption{Dataset statistics in ROBIN. Shown are the number of images per nuisance variable (top) and the number of images per object category (bottom). All images are annotated with class labels, bounding boxes and 3D poses.}
   % \setlength%\tabcolsep{0.5em}
    \begin{tabular}{l|c|c|c|c|c}
    \toprule
    \#Images     &  Shape   & 3D Pose  &  Texture & Context & Weather \\
    \midrule
    2632         &  563     & 517      &  535     &   444   &  573 \\
    \bottomrule
    \end{tabular}
    
    \begin{tabular}{c|c|c|c|c|c|c|c|c|c}
    \toprule
     Aeroplane & Bus & Car & Train & Boat & Bicycle & Motorbike & Chair & Diningtable & Sofa \\
    \midrule
    320 & 217 & 326 & 342 & 247 & 353 & 397 & 139 & 164 & 137 \\
    \bottomrule
    %\vspace{.01cm}
    \end{tabular}

    \label{tab:stats_per_cls}
%    \vspace{-0.7cm}
\end{table}

\end{comment}

\begin{figure*}[t]
    \centering
    \includegraphics[width=0.95\textwidth]{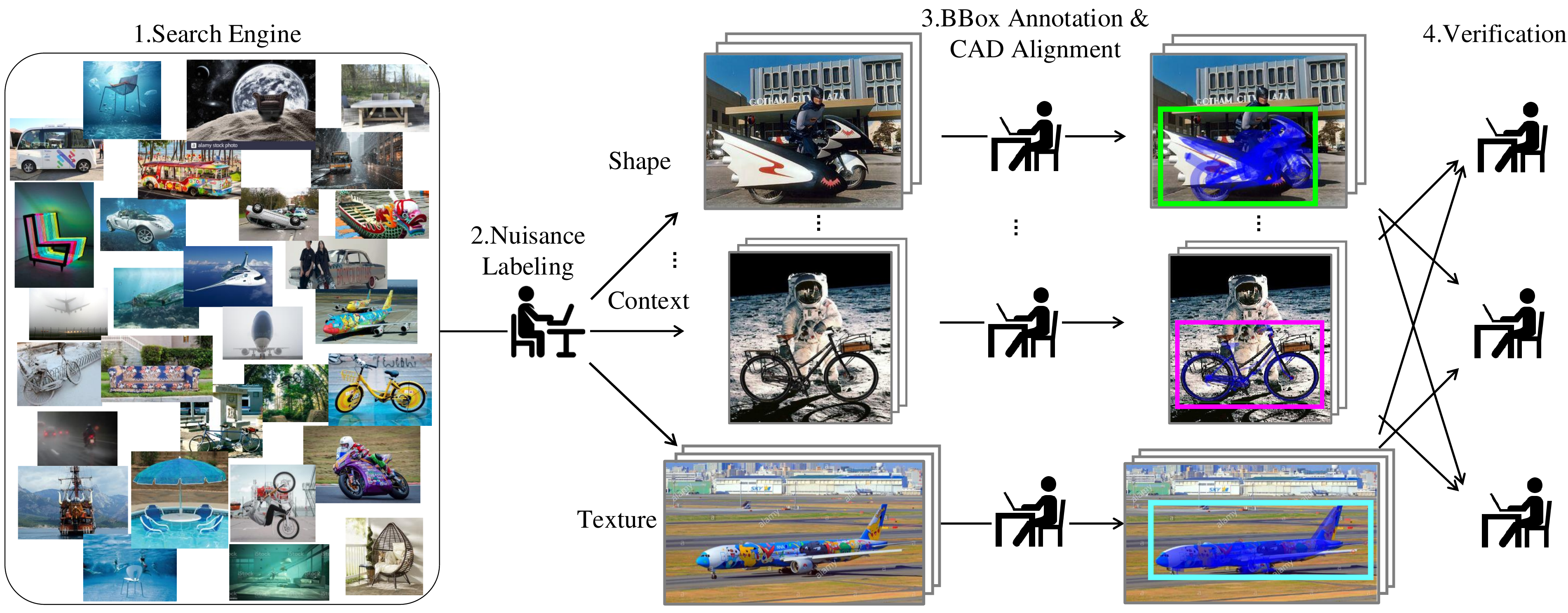}
    \caption{The data is collected from internet using a predefined set of search keywords, we then manually filter out all images that do not have OOD nuisances or have multiple nuisances. After collecting and splitting the data into different collections with different nuisance, we label images with object bounding boxes and align a CAD model to estimate the 3D pose of the object. The CAD models are overlaid on the images in blue. After each image annotation has been verified by at least two other annotators, we include it in our final dataset.}
    \label{fig:data_collection}
%   \vspace{-0.7cm}
\end{figure*}

% -------------------
\subsection{Data annotation}
\label{sec:annotation}
A schematic illustration of the annotation process is shown in ~\cref{fig:data_collection}.
After collecting the images from internet, we first classify the images according to OOD nuisance factor following the ontology discussed in~\cref{sec:ontology}. Subsequently, we annotate the images to enable benchmarking of a variety of vision tasks. In particular, we annotate the object class, 2D bounding box and 3D object pose. 
Note that we include the 3D pose, despite the large additional annotation effort compared to class labels and 2D bounding boxes, because we believe that extracting 3D information from images is an important computer vision task.

The annotation of the bounding boxes follows the coco format~\cite{lin2014microsoft}. We used a web-based annotation tool~\footnote{https://github.com/jsbroks/coco-annotator} that enables the data annotation with multiple annotators in parallel. 
The 3D pose annotation mainly follows the pipeline of PASCAL3D+~\cite{xiang_wacv14} and we use a slightly modified annotation tool from the one used in the PASCAL3D+ toolkit~\footnote{https://cvgl.stanford.edu/projects/pascal3d.html}. Specifically, to annotate the 3D pose each annotator selects a CAD model from the ones provided in PASCAL3D+, which best resembles the object in the input image. 
%such that the 3D object shape matches the best match of the objects in bounding boxes, and 
Subsequently, the annotator labels several keypoints to align the 6D pose of the CAD model to the object in input image.
%The annotation time can be significantly reduced by removing the annotation of 3D pose, but this will also reduce the value of the dataset, as understanding the 3D pose and shape of an object is also important for computer vision tasks.
After we have obtained annotations for the images, we count the distribution of number of images in each category and for categories with less images than average, we continue to collect additional images from internet for the minority categories.
Following this annotation process, we collected labels for all $2632$ images covering all nuisance factors.
Finally, the annotations produced by every annotator are verified by at least two other annotators to ensure the annotation is correct.
We have a total of $5$ annotators, and it took about $15$ minutes per image, resulting in more than $650$ hours of annotation effort.
%we defined previously, \cref{tab:number_images_category_bias} shows a statistics of the image numbers in our annotated dataset.

\noindent \textbf{Dataset splits.} 
To benchmark the IID performance, we split the $10665$ images that we retained from the PASCAL3D+ dataset into $8532$ training images and $2133$ test images.
%A test split without any nuisances is also created by select a set of images from the training set to evaluate the no nuisance baseline performance.
The OOD dataset splits for the nuisances "texture", "context", and "weather" can be directly used from our collected data.
%For nuisances like texture, context, and weather, the split is done via the sorting procedure, 
%for nuisances in object shape and pose, we use the original PASCAL3D+ annotations to filter the training set to ensure that the training set does not have any objects that have the same pose or shape with our curated test set.
%We split the data into different subsets that have different OOD nuisances against the training set.
As the Pascal3D+ data is highly variable in terms of 3D pose and shape, we create OOD splits w.r.t. the nuisances "pose" and "shape" by biasing the training data using the pose and shape annotations, such that the training and test set have no overlap in terms of shape and pose variations. These initial OOD splits are 
%In this way, we can also have an initial biased test set for testing shape and 3D pose nuisances using the original images from PASCAL3D+, these two test sets are 
further enhanced using the data we collected from the internet. %(Table \ref{tab:stats_per_cls}). 
The dataset and a detailed documentation of the dataset splits is available online\footnote{http://ood-cv.org/, Also see the supplementary material.}.

%Detailed statistics will be put in our supplementary materials.
%\begin{table*}[ht]
%\centering
%\caption{\label{tab:number_images_category_bias}\TODO{Caption, collect the numbers from the dataset}}
%\begin{tabular}{lrrrrrr}
%\toprule
%\#img       & Novel Shape  & Novel Pose    & Novel Texture   & Novel Context & Novel Weather & Total \\
%\midrule
%aeroplane   &  27         &   40         &  66             &  79          &   108          &   320    \\
%bus         &  83         &   18         &  82             &  4            &   30          &    217   \\
%car         &  159         &   24         &  40             &  20           &   83          &   326    \\
%train       &  34        &   42         &  130            &  70           &   66          &    342   \\
%boat        &  30         &   82         &  29             &  30           &   76          &   247    \\
%bicycle     &  64         &   70         &  28             &  78           &   113          &  353      \\
%motorbike   &  89         &   108         &  76             &  27           &   97           & 397      \\
%chair       &  40         &   40         &  42             &  17           &   0          &  139     \\
%diningtable &  22         &   65         &  18              &  59           &   0          & 164      \\
%sofa        &  15         &   28         &  24             &  60          &   0           &  127     \\
%\midrule
%Total       &  563         &   517         &  535      &  444    &   573          &   2632    \\
%\bottomrule
%\end{tabular}
%\end{table*}

%\vspace{-0.1cm}
\section{Experiments}
\label{sec:exp}
%\vspace{-0.10cm}
We test the robustness of vision models w.r.t. out-of-distribution shifts of individual nuisance factors in~\cref{sec:individual_nuisance} and evaluate popular methods for enhancing the model robustness of vision models using data augmentation techniques (\cref{sec:data_aug}) and changes to the model architecture (\cref{sec:model_architecture}).
%, to better understand the influence of nuisances, 
Finally, we study the effect when multiple nuisance factors are subject to OOD shifts in~\cref{sec:combined_nuisance} and give a comprehensive discussion of our results in~\cref{sec:summary_disc}.
%over vision models by further splitting the train set and the test set to not have shape or pose in common.

\noindent \textbf{Experimental Setup.}
%We study robustness of vision models in realistic out-of-distribution scenarios.
Our \OURS dataset enables benchmark vision models for three popular vision tasks: image classification, object detection, and 3D pose estimation.
We study robustness of popular methods for each task w.r.t. OOD shifts in five nuisance factors: object shape, 3D pose, object texture, background context and weather conditions.
%The ground truth labels for object detection and 3D pose estimation are annotated in the annotation process described in~\cref{sec:annotation}, 
We use the standard evaluation process of mAP@50 and Acc@$\frac{\pi}{6}$ for object detection and 3D pose estimation respectively.
For image classification, we crop the objects in the images based on their bounding boxes to create object-centric images, and use the commonly used Top-1 Accuracy to evaluate the performance of classifiers.
% TODO include detailed training recipe in supp
In all our experiments, we control variables such as the number of model parameters, model architecture, and training schedules to be comparable and only modify those variables we wish to study. The models for image classification are pre-trained on ImageNet \cite{deng2009imagenet} and fine-tuned on our benchmark. As datasets for a large-scale pre-training are not available for 3D pose estimation, we randomly initialize the pose estimation models and directly train them on the \OURS training split. Detailed training settings for vision models and data splits can be found in our supplementary.
%\TODO{describe the model architectures somewhere.}
%Unless otherwise stated, we use ResNet-50, Faster-RCNN with ResNet-50 backbone, and Res50-Specific for image classification, object detection, and 3D pose estimation respectively.

%\noindent \textbf{Implementation details}~~~
%We use 
%\begin{comment}
%In order to ensure a corrected evaluation of nuisance factors like object shape and pose, we split 10,665 images from the original Pascal3D+ dataset for training.
%To evaluate the performance of object detection, we follow the practice of Pascal-VOC~\cite{everingham2015pascal} dataset to computer the mAP-50 results using the prediction from the model and the ground truth labels.
%To evaluate the performance of image classification and 3D pose estimation, we cropped out the bounding boxes from the images to form a processed image dataset for classifier and pose estimation.
%\cref{fig:data_demo} shows examples of our annotated images.
%\end{comment}
% section about individual nuisances
% message: different vision tasks are influenced differently by different individual nuisances.
% ----------
\begin{table}
%\vspace{-0.7cm}
\small
\centering
\caption{\label{tab:individual_nuisance_on_different_tasks} Robustness to individual nuisances of popular vision models for different vision tasks. 
We report the performance on i.i.d. test data and OOD shifts in the object shape, 3D pose, texture, context and weather.
Note that image classification models are most affected by OOD shifts in the weather, while detection and pose estimation models mostly affected by OOD shifts in context and shape, suggesting that vision models for different tasks rely on different visual cues.}
\resizebox{\textwidth}{!}{
\begin{tabular}{c|lcccccc}
\toprule
Task  &                  & i.i.d             & shape             & pose              & texture          & context           & weather   \\ 
\midrule
\multirow{2}{*}{\begin{tabular}{c}Image\\Classification\end{tabular}} 
 & ResNet50              & 85.2\%$\pm$2.1\% & 73.2\%$\pm$1.9\%  & 73.8\%$\pm$2.0\%  & 76.2\%$\pm$2.6\% & 78.7\%$\pm$2.8\% & 69.2\%$\pm$1.9\% \\
 %                                           -12.0               -11.4               -9                 -6.5               -16
 & MbNetv3-L             & 81.5\%$\pm$1.7\% & 68.2\%$\pm$2.0\%  & 71.4\%$\pm$1.6\%  & 72.1\%$\pm$2.4\% & 75.9\%$\pm$2.9\% & 66.5\%$\pm$2.5\%   \\
\midrule
\multirow{2}{*}{\begin{tabular}{c}Object\\Detection\end{tabular}}
 & Faster-RCNN           & 72.6\%$\pm$1.7\% & 61.6\%$\pm$2.4\%  & 62.4\%$\pm$1.7\%  & 56.3\%$\pm$1.1\% & 35.6\%$\pm$1.8\% & 50.7\%$\pm$1.6\% \\ 
 %                                           -11                 -10.2               -16.3              -37                -21.9
 & RetinaNet             & 74.7\%$\pm$1.6\% & 64.1\%$\pm$2.0\%  & 65.8\%$\pm$1.9\%  & 61.5\%$\pm$2.0\% & 40.3\%$\pm$2.2\% & 54.2\%$\pm$2.0\% \\ 
\midrule
\multirow{2}{*}{\begin{tabular}{c}3D Pose\\Estimation\end{tabular}}
 & Res50-Specific         & 62.4\%$\pm$2.4\% & 43.5\%$\pm$2.5\%  & 45.2\%$\pm$2.8\%  & 51.4\%$\pm$1.8\% & 50.8\%$\pm$1.9\% & 49.5\%$\pm$2.1\% \\ 
 %                                           -18.9               -17.2               -11                -11.6              -12.9
 & NeMo                  & 66.7\%$\pm$2.3\% & 51.7\%$\pm$2.3\%  & 56.9\%$\pm$2.7\%  & 52.6\%$\pm$2.0\% & 51.3\%$\pm$1.5\% & 49.8\%$\pm$2.0\% \\ 
\bottomrule
\end{tabular}}

%\vspace{-0.7cm}
\end{table}

\subsection{Robustness to individual nuisances}
\label{sec:individual_nuisance}
%\noindent \textbf{OOD nuisances have different effect on vision models for different visual tasks.}~~~
The \OURS benchmarks enables, for the first time, to study the influence of OOD shifts in individual nuisance factors on tasks of classification, detection and pose estimation.
We first study the robustness of one representative methods for each task. 
In~\cref{tab:individual_nuisance_on_different_tasks}, we report the test performance on a test set with i.i.d. data, as well as the performance under OOD shifts to all five nuisance factors that are annotated in the \OURS benchmark.
We observe that for image classification, the performance of the classic ResNet50 architecture~\cite{he2015deep} drops significantly for every OOD shift in the data. The largest drop is observed under OOD shifts in the weather conditions ($-16.0\%$), while the performance drop for OOD context is only $-6.5\%$. 
The results suggests that the model does not rely very much on contextual cues but rather focuses more on the overall gist of the image, which is largely affected by changing weather conditions. Moreover, the classification model is more affected by OOD shifts in geometric cues such as the shape and the 3D pose, compared to the object texture.
On the contrary, for object detection the performance of a Faster-RCNN~\cite{ren2015faster} model drops the most under OOD context ($-37\%$ mAP@50), showing that detection models rely strongly on contextual cues. 
While the performance of the detection model also decreases significantly across all OOD shifts, the appearance-based shifts like texture, context and weather have a stronger influence compared to OOD shifts in the shape and pose of the object. 
For the task of 3D pose estimation, we study a ResNet50-Specific~\cite{zhou2018starmap} model, which is a common pose estimation baseline that treats pose estimation as a classification problem (discretizing the pose space and then classifying an image into one of the pose bins).
We observe that the performance for 3D pose estimation drops significantly, across all nuisance variables and most prominently for OOD shifts in the shape and pose.

\begin{figure}
     \centering
     \begin{subfigure}[b]{0.3\textwidth}
         \centering
         \includegraphics[width=\textwidth]{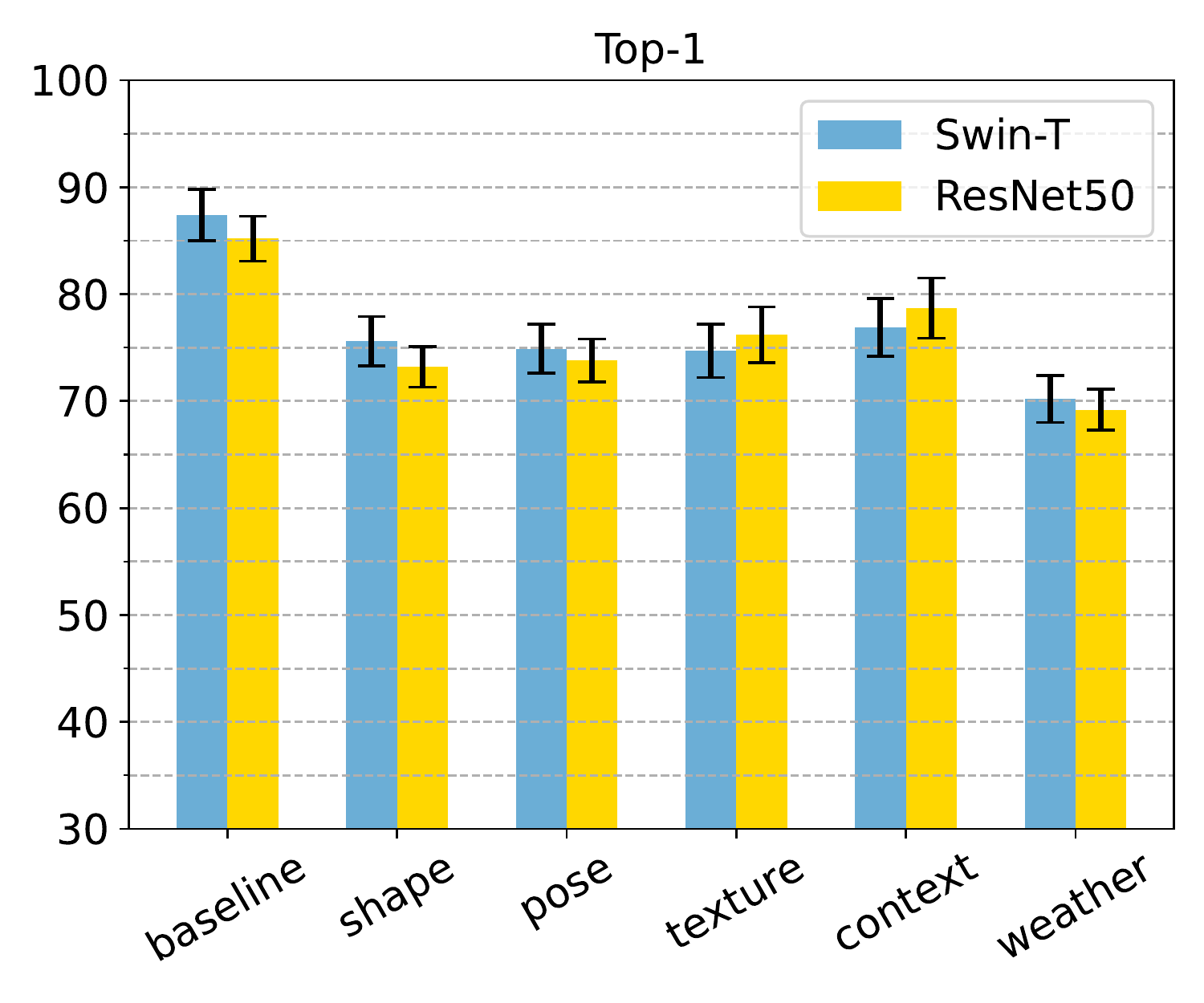}
         \caption{Image Classification}
         \label{fig:cnn_vs_transformer_classification}
     \end{subfigure}
     \begin{subfigure}[b]{0.3\textwidth}
         \centering
         \includegraphics[width=\textwidth]{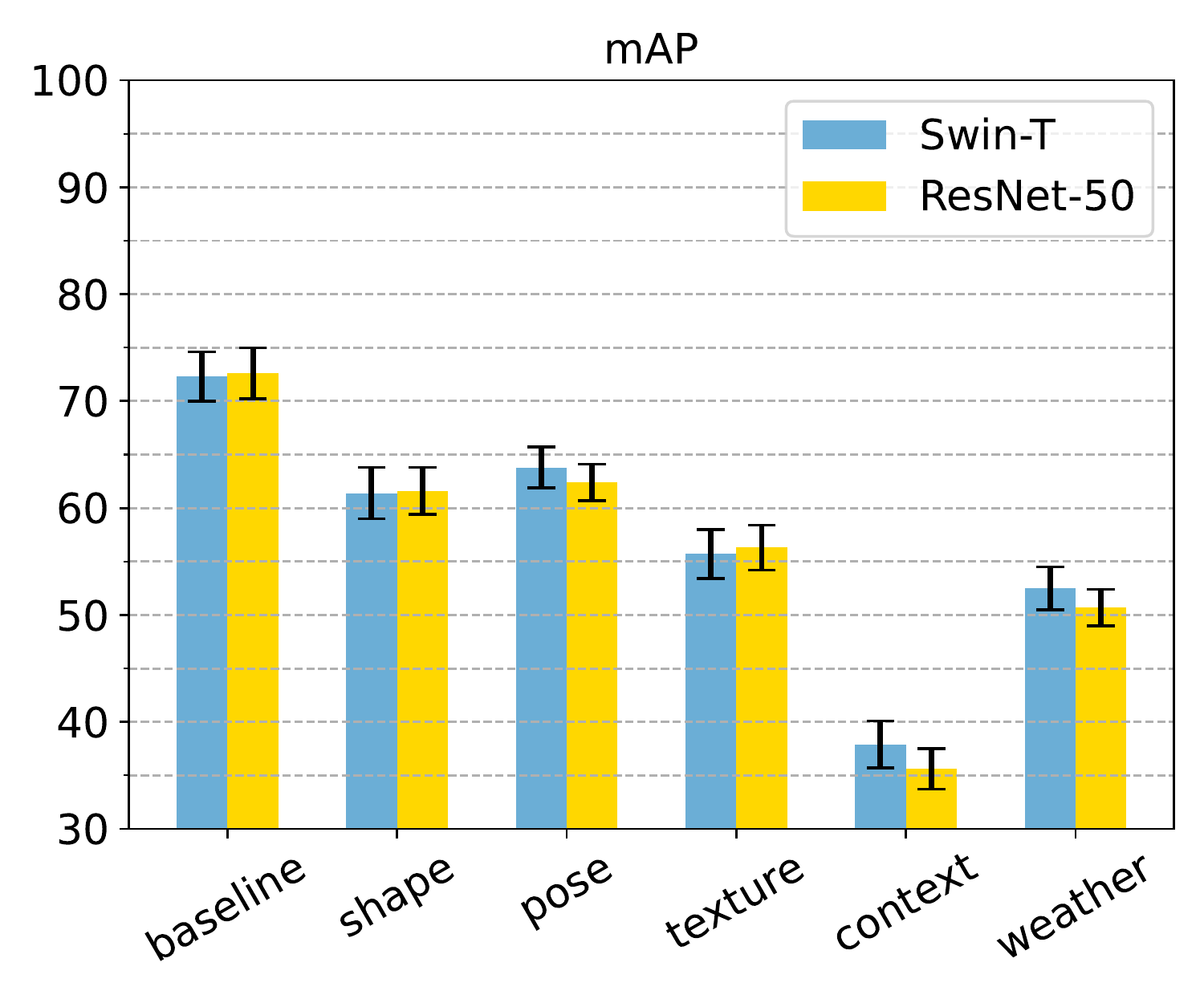}
         \caption{Object Detection}
         \label{fig:cnn_vs_transformer_detection}
     \end{subfigure}
     \begin{subfigure}[b]{0.3\textwidth}
         \centering
         \includegraphics[width=\textwidth]{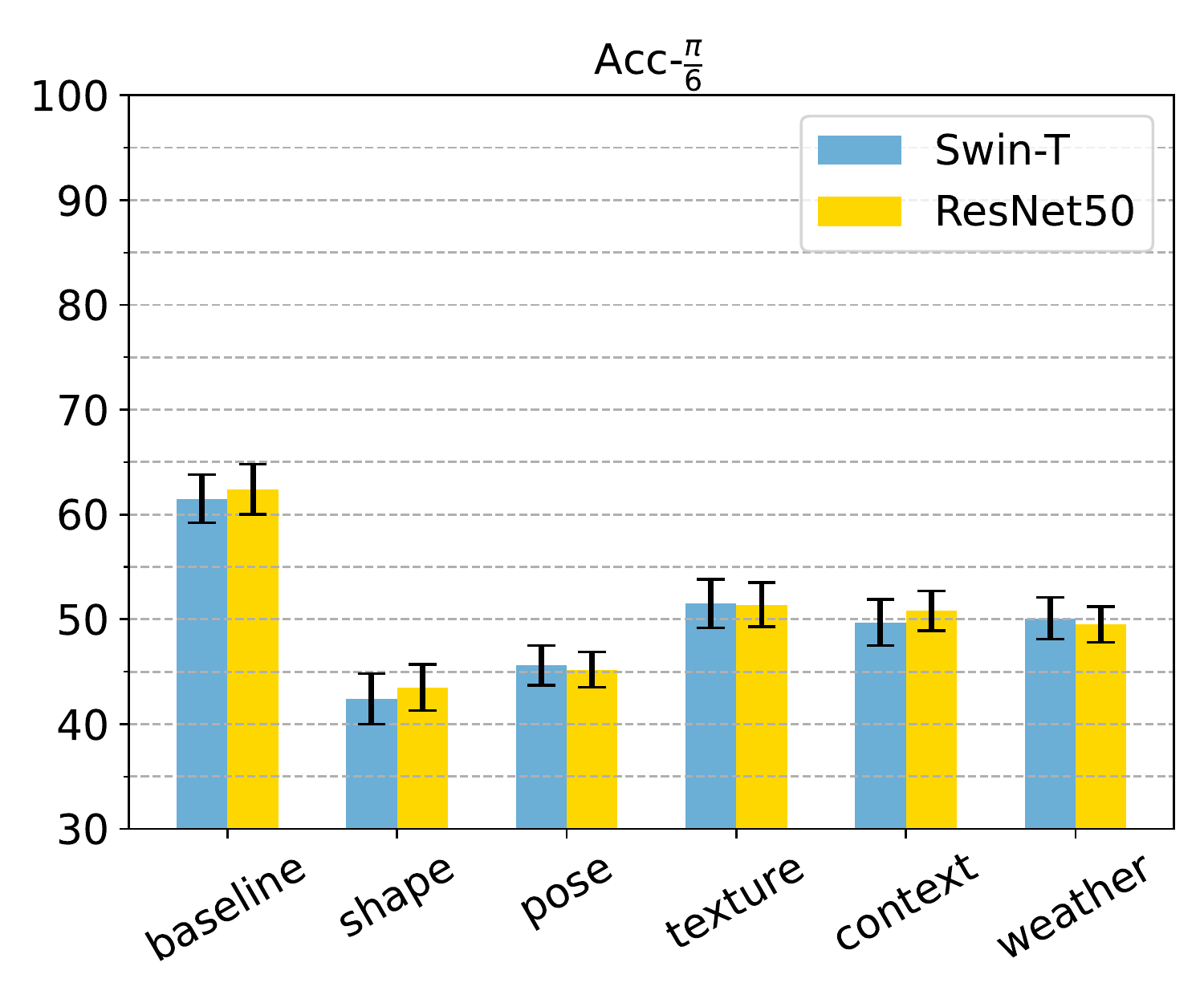}
         \caption{3D Pose Estimation}
         \label{fig:cnn_vs_transformer_estimation}
     \end{subfigure}
        \caption{Performance of CNN and Transformer on our benchmark. Transformers have a higher in-domain performance, but CNNs and transformers degrades mostly the same on OOD testing examples.}
        \label{fig:cnn_vit}
        
%\vspace{-0.5cm}
\end{figure}
%\setlength{\belowdisplayskip}{3pt}
%\vspace{-0.08cm}

In summary, our experimental results show that \textbf{OOD nuisances have different effect on vision models for different visual tasks}. 
This suggests OOD robustness should not be simply treated as a domain transfer problem between datasets, but instead it is important to study the effects of individual nuisance factors. 
Moreover, OOD robustness might require different approaches for each vision tasks, as we observe clear differences in the effect of OOD shifts in individual nuisance factors between vision tasks.

%#####################################################
\begin{table}[t]
\small
\caption{Effect of data augmentation techniques on OOD robustness for three vision tasks. We report the performance of one baseline model for each task, as well as the same model trained with different augmentation techniques: Stylizing , AugMix\cite{hendrycks2019augmix} and Adversarial Training \cite{wong2020fast}.
We evaluate all models on i.i.d. test data and OOD shifts in the object shape, 3D pose, texture, context and weather. 
Strong data augmentation only improves robustness to appearance-based nuisances but even decreases the performance to geometry-based nuisances like shape and 3D pose. }
\begin{subtable}{\textwidth}
\centering

\begin{tabular}{lcccccc}
\toprule
top-1                             & i.i.d & shape & pose &  texture & context & weather \\ 
\midrule
ResNet-50                          & 85.2\% & 73.2\% & 73.8\% & 76.2\% & 78.7\% & 69.2\% \\
Style Transfer                    & 86.4\% & 72.8\% & 72.6\% & 78.9\% & 78.8\% & 73.6\% \\ 
AugMix & 87.6\% & 73.0\% & 73.3\% & 82.1\% & 82.6\% & 75.2\% \\
%                                           -0.2     -0.5     +5.9     +3.9     +6
Adv. Training                     & 83.7\% & 72.1\% & 71.6\% & 77.9\% & 79.9\% & 72.6\% \\
\bottomrule
\end{tabular}
\caption{Top-1 accuracy results on image classification}
\label{tab:data_aug_cls}
\end{subtable}

\begin{subtable}{\textwidth}
\centering
\begin{tabular}{lcccccc}
\toprule
mAP-50                            & i.i.d   & shape  & pose   & texture & context & weather \\ 
\midrule
Faster-RCNN                          & 72.6\% & 61.6\% & 62.4\% & 56.3\% & 35.6\% & 50.7\%  \\
% AugMix is not direct applicable to detection
Style Transfer                    & 73.1\% & 59.8\% & 61.3\% & 58.4\% & 39.4\% & 53.5\% \\ 
Adv. Training                     & 71.3\% & 60.4\% & 60.1\% & 57.4\% & 36.9\% & 52.8\% \\
\bottomrule
\end{tabular}
\caption{mAP@50 results on object detection}
\label{tab:data_aug_det}
\end{subtable}

\begin{subtable}{\textwidth}
\centering
\begin{tabular}{lcccccc}
\toprule
Acc-$\frac{\pi}{6}$               & i.i.d   & shape  & pose   & texture & context & weather \\ 
\midrule
Res50-Spec.                         & 62.4\% & 43.5\% & 45.2\% & 51.4\% & 50.8\% & 49.5\%  \\
Style Transfer                    & 63.1\% & 41.8\% & 44.7\% & 55.8\% & 54.3\% & 53.8\% \\ 
AugMix & 64.8\% & 44.1\% & 44.8\% & 56.7\% & 54.7\% & 55.6\% \\
Adv. Training                     & 61.1\% & 41.7\% & 43.5\% & 52.4\% & 51.7\% & 50.9\% \\
\bottomrule
\end{tabular}
\caption{Acc-$\frac{\pi}{6}$ results on pose estimation}
\label{tab:data_aug_pose}
\end{subtable}

\label{tab:data_aug}
%\vspace{-1cm}
\end{table}

%\vspace{-0.1cm}
\subsection{Data Augmentation for Enhancing Robustness}
\label{sec:data_aug}
%\vspace{-0.1cm}
%We first test the effectiveness of data augmentation techniques for improving the robustness.
%\noindent \textbf{Data augmentation techniques can only help with appearance-based nuisances.}~~~
Data augmentation techniques have been widely adopted as an effective means of improving the robustness of vision models. Among such data augmentation methods, stylizing images with artistic textures~\cite{geirhos2018}, mixing up the original image with a strongly augmented image (AugMix~\cite{hendrycks2019augmix}), and adversarial training~\cite{wong2020fast} are the most effective methods.
We test these data augmentation methods on \OURS to find out if and how they affect the OOD robustness.
The experimental results are summarized in~\cref{tab:data_aug}.
Overall, AugMix~\cite{hendrycks2019augmix} improves the OOD robustness the most for image classification and pose estimation. While AugMix is not directly applicable to object detection, we observe that strong data augmentation style transfer~\cite{gatys2016image} leads to a better improvement compared to adversarial training.
Importantly, these data augmentation methods improve the OOD robustness mostly w.r.t. appearance-based nuisances like texture, context, and weather.%, \eg, AugMix improve the top-1 accuracy of the classification model by $5.9\%$, $3.9\%$, and $6\%$ on novel texture, context, and weather respectively.
However, in all our experiments \textit{data augmentation slightly reduces the performance} under OOD shape and 3D pose.
%, \eg, the performance of AugMix for shape and pose decreases by $0.2\%$ and $0.5\%$ top-1 accuracy on a classification model.
We suspect that this happens because data augmentation techniques mostly change appearance-based properties of the image and do not change the geometric properties of the object (i.e. shape and 3D pose).
%3D objects in the image can not be directly operated by current data augmentation techniques which only operates on the pixels, therefore data augmentation can not help increasing the robustness of the model for images containing objects with geometric-based nuisances like shape and pose.
Similar trends are observed across all three of the tasks we tested, image classification, object detection, and pose estimation.
These results suggest that two categories of nuisances exists, namely \textit{appearance-based} nuisances like novel texture, context, and weather, and \textit{geometric-based} nuisances like novel shape and pose.
We observe that \textbf{data augmentation only improves robustness of appearance-based nuisances but can even decrease the performance w.r.t. geometry-based nuisances}.

%\vspace{-0.2cm}
\subsection{Effect of Model Architecture on Robustness}
\label{sec:model_architecture}
%\vspace{-0.1cm}
%Another promising way of improving the robustness is to change the architecture of the model.
In this section, we investigate four popular architectural changes that have proven to be useful in real world applications. Paricularly, we evaluate \textit{CNNs vs Transformers}, the \textit{model capacity}, \textit{one stage vs two stage} detectors, and models with \textit{integrated 3D priors}. 
Note that when we change the model architecture we keep other parameters such as number of parameters and capacity the same.
% we keep all the other variants controlled except the variants we are interested in.

%\vspace{-0.7cm}
\begin{table}[t]
\small
\centering
\caption{\label{tab:mbv3_vs_r50} OOD robustness of models with different capacities. While the performance degradation of MobileNetv3-Large (MbNetv3-L) are about the same as those of ResNet-50, training with data augmentation technique has smaller effect on MbNetv3-L due to the limited capacity.}
\setlength\tabcolsep{0.3em}
\begin{tabular}{lcccccc}
\toprule
                                   & i.i.d   & shape  & pose   & texture   & context  & weather   \\ 
\midrule
ResNet50                           & 85.2\% & 73.2\% & 73.8\% & 76.2\%    & 78.7\%   & 69.2\% \\
%                               degradation  -12      -11.4    -9          -6.5       -16
+AugMix~\cite{hendrycks2019augmix} & 87.6\% & 73.0\% & 73.3\% & 82.1\%    & 82.6\%   & 75.2\% \\
%                               improvement  -0.2     -0.5     +5.9        +3.9       +6
\midrule
MbNetv3-L~\cite{howard2019searching_mbv3}   & 81.5\%  & 68.2\%  &  71.4\%  & 72.1\% & 75.9\% & 66.5\% \\ 
%                                         degradation  -13.3      -10.1     -9.4     -5.6     -15
+AugMix~\cite{hendrycks2019augmix}          & 83.1\%  & 67.8\%  &  71.6\%  & 74.3\% & 76.8\% & 69.7\%  \\
%                                         improvement  -0.4       +0.2      +2.2     +0.9     +3.2
\bottomrule
\end{tabular}

%\vspace{-0.5cm}
\end{table}

\noindent \textit{CNNs vs Transformers.}~~
%\noindent \textbf{The robustness of Transformers and CNNs against individual real-world nuisances are about the same.}~~~
Transformers have emerged as a promising alternative to convolutional neural networks (CNNs) as an architecture for computer vision tasks recently~\cite{dosovitskiy2020image,liu2021swin}. 
While CNNs have been extensively studied for robustness, the robustness of vision transformers are still under-explored.
Some works~\cite{bhojanapalli2021understanding,mahmood2021robustness} have shown that transformer architecture maybe more robust to adversarial examples, but it remains if this result holds for OOD robustness.
%With our proposed benchmark, we are able to compare the robustness of both CNNs and vision transformers in more depth.
%Specifically, with the separated nuisance factors in our dataset, we can compare if the performance degradation of CNNs and vision transformers are the same on different individual nuisance factors.
In the following, we compare the performance of CNNs and transformers on the tasks of image classification, object detection and 3D pose estimation on the \OURS benchmark.
Specifically, we replace the backbone the vision models for each task from ResNet-50 to Swin-T~\cite{liu2021swin}.% to ensure that other than the architectural difference of transformer and CNNs, the other variable are roughly the same.
% And how does different data augmentations techniques in~\cref{sec:data_aug} influences the performance of CNNs and transformers. To simplify the comparison process. % TODO update this.
Our experimental results are presented in~\cref{fig:cnn_vit}. Each experiment is performed five times and we report mean performance and standard deviation.
It can be observed that CNNs and vision transformers have a comparable performance across all tasks as the difference between their performances are within the margin of error. 
Particularly, we do not observe any enhanced robustness as OOD shifts in individual nuisance factors lead to a similar decrease in performance in both the transformer and the CNN architecture.
While we observe a slight performance gain on i.i.d. data in image classification (as reported in many other works), our results suggest that \textbf{Transformers do not have any enhanced OOD robustness compared to CNNs}.
Note our findings here contrast with previous work on this topic~\cite{bai2020vitsVScnns}, we argue that this is because our benchmark enables the study for individual nuisance factors on real world images, and the control over different individual nuisances give us opportunity to observe more errors in current vision models.

% message here, CNN and transformers are mostly the same, or maybe transformer is more biased towards shape and CNNs bias towards texture.
% how does data augmentation influences this

% MobileNetV3 vs resnet50
% message here:
% the nuisances have mostly equal influence on both small model and larger models, but it is harder to use data augmentations to improve the robustness of small models.
\noindent \textit{Model capacity.}~~~
%\noindent \textbf{Due to the limited capacity, it is hard to improve the robustness of efficient models.}~~~
For deployment in real applications, smaller models are preferred because they can yield better efficiency than regular models.
In the following, we compare image classification performance of MobileNetV3~\cite{howard2019searching_mbv3} in~\cref{tab:mbv3_vs_r50}.
Compared to ResNet-50, MobileNetv3 suffers a similar performance degradation under OOD shifts in the data.
However, data augmentations does not improve the robustness of MobileNetV3~\cite{howard2019searching_mbv3} as much as for ResNet-50, \eg, performance on context nuisances improved by $3.9\%$ for ResNet-50, but the improvement is only $0.9\%$ for MobileNetV3.  
This suggests that \textbf{OOD robustness is more difficult to achieve for efficient models with a limaited capacity}.

%\vspace{-0.8cm}
\begin{table}[t]
\footnotesize
\centering
\caption{\label{tab:one_two_stage} Comparison between one-stage method and two-stage object detection methods. One-stage methods are more robust compared to two-stage methods.}
\setlength\tabcolsep{0.3em}
\begin{tabular}{lcccccc}
\toprule
                                     & i.i.d   & shape  & pose   & texture & context & weather   \\ 
\midrule
RetinaNet~\cite{lin2017focal}        & 74.7\% & 64.1\% & 65.8\% & 61.5\%  & 40.3\%  & 54.2\% \\
%                                              -10.6    -8.9     -13.2     -34.4     -20.5
+Style Transfer~\cite{geirhos2018}   & 75.8\% & 62.7\% & 64.2\% & 63.7\%  & 44.7\%  & 55.8\% \\
\midrule
Faster-RCNN~\cite{ren2015faster}     & 72.6\% & 61.6\% & 62.4\% & 56.3\%  & 35.6\%  & 50.7\%  \\
%                                              -11      -10.2    -16.3     -37       -21.9
+Style Transfer~\cite{geirhos2018}   & 73.1\% & 59.8\% & 61.3\% & 58.4\%  & 39.4\%  & 53.5\% \\ 
\bottomrule
\end{tabular}

%\vspace{-1.5cm}
\end{table}

\begin{table}[t]
\small
\centering
\caption{\label{tab:pose_nemo_r50}Robustness of 3D pose estimation methods. 
We compare ``Res50-Specific'', which treats pose estimation as classification problem, and ``NeMo'', which represents the 3D object geometry explicitly.
We observe OOD shifts in shape and pose leads to more performance degradation. NeMo has a significantly enhanced performance to OOD shifts in object shape and pose.}
\setlength\tabcolsep{0.3em}
\begin{tabular}{lcccccc}
\toprule
                                     & i.i.d   & shape  & pose   & texture & context & weather   \\ 
\midrule
Res50-Specific                     & 62.4\% & 43.5\% & 45.2\% & 51.4\% & 50.8\% & 49.5\%  \\
+AugMix~\cite{hendrycks2019augmix} & 64.8\% & 44.1\% & 44.8\% & 56.7\% & 54.7\% & 55.6\% \\
\midrule
NeMo~\cite{wang2021nemo}           & 66.7\% & 51.7\% & 56.9\% & 52.6\% & 51.3\% & 49.8\%  \\
+AugMix~\cite{hendrycks2019augmix} & 67.9\% & 53.1\% & 58.6\% & 57.8\% & 55.1\% & 56.7\% \\
\bottomrule
\end{tabular}
%\vspace{-.5cm}
\end{table}

% Retinanet vs Faster rcnn
% message here, one stage and two stage are basically the same.
\noindent \textit{One stage vs two stage for detection.}~~~
%\noindent \textbf{Two stage detection methods are not necessarily more robust than one-stage methods.}~~~
It is a common belief in object detection community that two-stage detectors are more accurate, while one-stage detectors are more efficient.
For object detection task, two popular types of architecture exist, namely one-stage and two stage models.
We tested two representative models from these architecture types, RetinaNet~\cite{lin2017focal}, a one-stage detector, and Faster-RCNN~\cite{ren2015faster}, which is a two-stage detector.
From our results in~\cref{tab:one_two_stage}, we observe that RetinaNet achieves a higher performance compared to Faster-RCNN on the \OURS benchmark.
However, when accounting for improved i.i.d performance, the OOD performance degradation are similar between two models.
These initial result suggests that \textbf{two-stage methods achieve a higher score than one-stage methods, but are not necessarily more robustness}. % further investigation on this topic is needed.

% NeMo vs ResNet-specific
\noindent \textit{Models with explicit 3D object geomtery.}~~~
%\noindent \textbf{Integrating 3D priors into models help with geometric-based nuisances.}~~~
Recently, Wang et al.~\cite{wang2021nemo} introduced NeMo, a neural network architecture for 3D pose estimation that explicitly models 3D geometery, and they demonstrated promising results on enhancing robustness to partial occlusion and unseen 3D poses.
In~\cref{tab:pose_nemo_r50}, we compare NeMo~\cite{wang2021nemo} model and a general Res50-Specific model on task of pose estimation on \OURS benchmark.
NeMo~\cite{wang2021nemo} shows a stronger robustness against geometric-based nuisances (shape and pose), while robustness on appearance-based nuisances is comparable.
This result suggests that, \textbf{neural networks with an explicit 3D object representation have a largely enhanced robustness to OOD shifts in geometry-based nuisances}. 
These results seem complementary to our experiments in the previous section, which demonstrate that strong data augmentation can help to improve the robustness of vision models to appearance-based nuisances, but not to geometry-based nuisances.

We further investigate, if robustness against all nuisance types can be improved by combining data augmentation with architectures that explicitly represent the 3D object geometry. Specifically, we train NeMo~\cite{wang2021nemo} with strong augmentations like AugMix~\cite{hendrycks2019augmix} and our results in~\cref{tab:pose_nemo_r50} show that this indeed largely enhances the robustness to OOD shifts in appearance-based nuisances, while retaining (and slightly improving) the robustness to geometry-based nuisances.
Result suggests that enhancements of robustness to geometry-based nuisances can be developed independently to those for appearance-based nuisances.
% message here:
% cnn vs transformer, transformer is not better than CNNs. (ResNet50 vs Swin-T)
% model size: light weight model vs large model.
% model type for det, one-stage vs two-stage.
% architectural changes, NeMo helps with novel pose.
%One of the unique characteristics about our proposed dataset is the separation of individual nuisance, from the experimental results presented in previous sections, we can observe that the influence of geometrical novelties like novel shape and novel pose for vision models are largely different from the influences of photometrical novelties like novel texture, novel context, and novel weather: photometrical novelties can be relative easily overcomed by using strong data augmentations.

\subsection{OOD shifts in Multiple Nuisances}
\label{sec:combined_nuisance}
%\noindent \textbf{Appearance-based nuisance and geometric-based nuisance amplifies each other when combined.}~~~
%Our dataset have separated nuisances factor labelled in our datasets, and in our previous experiments of benchmarking data augmentations techniques and model architectural changes, 
In our experiments, we observed that geometry-based nuisances have different effects compared to appearance-based nuisances.
In the following, we test the effect when OOD shifts happen in both of these nuisance types. 
Specifically, we introduce new dataset splits, which combine appearance-based nuisances, including texture, context, or weather, with the geometry-based nuisances shape and pose. From~\cref{tab:shape_pose_plus_x}, we observe \textbf{OOD shifts in multiple nuisances amplify each other}.
For example, for image classification, an OOD shift in only the 3D pose reduces the performance by $11.4\%$ from $85.2\%$ to $73.8\%$, and an OOD shift in the context reduces the performance by $6.6\%$.
However, when pose and context are combined the performance reduces by $24.5\%$. 
We observe a similar amplification behaviour across all three tasks, suggesting that it is a general effect that is likely more difficult to address compared to single OOD shifts.
%using data augmentation techniques like style transfer~\cite{geirhos2018} on these images with combined nuisances still helps, but the gain from augmentation are not as large as when the images has one single nuisance, also indicating that combined nuisances are harder to solve than just single nuisances.

% message here:
% combined biases will hurt the performance larger than individual bias
% using data augmentation (effective for textures etc) alone can not help with combined biases.

\begin{table}[t]
\centering
\caption{\label{tab:shape_pose_plus_x} Robustness to OOD shifts in multiple nuisances. When combined, OOD shifts in appearance-based nuisances and geometric-based nuisances amplifies each other, leads to further decrease compared to effects in individual nuisances.}
%\caption{\label{tab:shape_pose_plus_x} \todo{Explain, what models are used? what does the table show, what should be noted by the reader?} combined nuisance experiments. \todo{nuisances do not simply add up, e.g. in classification pose reduces by 12 and context by 7 but combined they reduce by 25.}}
\begin{tabular}{lcccc}
\toprule
                & i.i.d             & texture   & context       & weather   \\ 
\midrule
Classification  & 85.2\%           & 76.2\%    & 78.7\%        & 69.2\%      \\
%+style transfer &                  & 78.9\%    & 78.8\%        & 73.6\% \\ 
%\midrule
+ shape    & 73.2\%           & 62.8\%    & 63.6\%        & 51.2\%          \\
%+style transfer & 72.8\%           & 63.9\%    & 64.8\%        & 52.7\%          \\
%\midrule
+ pose     & 73.8\%           & 61.9\%    & 60.7\%        & 49.8\%          \\
%+style transfer & 72.6\%           & 62.8\%    & 62.6\%        & 50.4\%          \\
\midrule
Detection       & 72.6\%           & 56.3\%    & 35.6\%        & 50.7\%         \\
%+style transfer &                  & 58.4\%    & 39.4\%        & 53.5\%          \\
%\midrule
+ shape    & 61.6\%           & 41.2\%    & 24.3\%        & 30.7\%       \\
%+style transfer & 59.8\%           & 42.5\%    & 27.6\%        & 34.1\%        \\
%\midrule
+ pose     & 62.4\%           & 45.6\%    & 26.1\%        & 29.8\%        \\
%+style transfer & 61.3\%           & 46.7\%    & 27.4\%        & 33.9\%          \\
\midrule
Pose estimation & 62.4\%           & 51.4\%    & 50.8\%        & 49.5\%  \\ 
%+style transfer &                  & 56.1\%    & 54.3\%        & 55.6\% \\ 
%\midrule
+ shape    & 43.5\%           & 33.1\%    & 31.0\%        & 29.8\%        \\
%+style transfer & 44.1\%           & 38.2\%    & 39.7\%        & 36.8\%          \\
%\midrule
+ pose     & 45.2\%           & 30.2\%    & 29.7\%        & 28.1\%        \\
%+style transfer & 44.8\%           & 32.1\%    & 33.8\%        & 32.3\%          \\
\bottomrule
\end{tabular}

%than simply adding up different nuisances, \eg, the nuisances in texture reduces the performance of detection models by $16.3\%$, and nuisances in shape reduce by $11\%$, and the combined nuisances of shape+texture reduce the performance by $31.4\%$.}
%\vspace{-0.5cm}
\end{table}

\section{Conclusion}
\label{sec:summary_disc}
% we have individual nuisances
% first we test how individual nuisances influence each vision tasks we tested.
% we found that different visual tasks degrade different to individual nuisances. (detection degrade on context, and pose estimation on shape and pose)
% we tested commonly used method to improve robustness, i) data augmentation techniques, ii) changes in model architecture
% we can classify the individual nuisance into two groups, namely the appearance nuisances and geometric nuisances.
% we found that, data augmentations can help with the appearance nuisance, but not the geometric nuisances.
% and by integrating 3D priors into the model, we found that the robustness to geometric nuisances can be improved, however, other promising architectural improvement such as Transformers over CNNs does not show advantages in terms of robustness against nuisances.
% solutions? :
% Compositional Representations, 3D representations, causality (lecun, hinton, bengio)

We have shown that proposed \OURS benchmark enables a thorough diagnosis of robustness of vision models to realistic OOD shifts in individual nuisance factors.
Overall, we observe that OOD shifts poses a great challenge to current state-of-the-art vision models and requires significant attention from the research community to be resolved.
Notably, we found that nuisance factors have a different effect on different vision tasks, suggesting that we might need different solutions for enhancing the OOD robustness for different vision tasks.
In our experiments, it can also be clearly observed that the nuisances can be roughly separated into two categories, \textit{appearance-based nuisances} like texture, context, or weather, and another one is \textit{geometry-based nuisances} such as shape or pose. 
We showed that strong data augmentation enhances the robustness against appearance-based nuisances, but has very little effect on geometric-based nuisances. 
On the other hand, neural network architectures with an explicit 3D object representation achieve an enhanced robustness against geometric-based nuisances.
While we observe that OOD robustness is largely an unsolved and severe problem for computer vision models, our results also suggest a way forward to address OOD robustness in the future. Particularly, that approaches to enhance the robustness may need to be specifically designed for each vision tasks, as different vision tasks focus on different visual cues.
Moreover, we observed a promising way forward to a largely enhanced OOD robustness is to develop neural network architectures that represent the 3D object geometry explicitly and are trained with strong data augmentation to address OOD shifts in both geometry-based and appearance-based nuisances combined.

\noindent \textbf{Acknowledgements.} AK acknowledges support via his Emmy Noether Research Group funded by the German Science Foundation (DFG) under Grant No. 468670075.
BZ acknowledges compute support from LunarAI.
AY acknowledges grants ONR N00014-20-1-2206 and ONR N00014-21-1-2812.

%%%%%%%%% REFERENCES
{\small
\bibliographystyle{ieee_fullname}
\bibliography{eccv_egbib}

\begin{thebibliography}{10}\itemsep=-1pt

\bibitem{rovi}
{Robust Vision Challenge 2020}.
\newblock \url{http://www.robustvision.net/}.

\bibitem{alcorn2019strike}
Michael~A Alcorn, Qi Li, Zhitao Gong, Chengfei Wang, Long Mai, Wei-Shinn Ku,
  and Anh Nguyen.
\newblock Strike (with) a pose: Neural networks are easily fooled by strange
  poses of familiar objects.
\newblock In {\em Proceedings of the IEEE Conference on Computer Vision and
  Pattern Recognition}, pages 4845--4854, 2019.

\bibitem{bai2020vitsVScnns}
Yutong Bai, Jieru Mei, Alan Yuille, and Cihang Xie.
\newblock Are transformers more robust than cnns?
\newblock In {\em Adv. Neural Inform. Process. Syst.}, 2021.

\bibitem{bengio2021deep}
Yoshua Bengio, Yann Lecun, and Geoffrey Hinton.
\newblock Deep learning for ai.
\newblock {\em Communications of the ACM}, 2021.

\bibitem{bhojanapalli2021understanding}
Srinadh Bhojanapalli, Ayan Chakrabarti, Daniel Glasner, Daliang Li, Thomas
  Unterthiner, and Andreas Veit.
\newblock Understanding robustness of transformers for image classification.
\newblock In {\em Int. Conf. Comput. Vis.}, 2021.

\bibitem{ilab}
Ali Borji, Saeed Izadi, and Laurent Itti.
\newblock ilab-20m: A large-scale controlled object dataset to investigate deep
  learning.
\newblock In {\em IEEE Conf. Comput. Vis. Pattern Recog.}, 2016.

\bibitem{chen2014detect}
Xianjie Chen, Roozbeh Mottaghi, Xiaobai Liu, Sanja Fidler, Raquel Urtasun, and
  Alan Yuille.
\newblock Detect what you can: Detecting and representing objects using
  holistic models and body parts.
\newblock In {\em IEEE Conf. Comput. Vis. Pattern Recog.}, 2014.

\bibitem{cubuk2018autoaugment}
Ekin~D Cubuk, Barret Zoph, Dandelion Mane, Vijay Vasudevan, and Quoc~V Le.
\newblock Autoaugment: Learning augmentation policies from data.
\newblock In {\em IEEE Conf. Comput. Vis. Pattern Recog.}, 2018.

\bibitem{cui2022discriminabilitytransferability}
Quan Cui, Bingchen Zhao, Zhao-Min Chen, Borui Zhao, Renjie Song, Jiajun Liang,
  Boyan Zhou, and Osamu Yoshie.
\newblock Discriminability-transferability trade-off: An information-theoretic
  perspective.
\newblock In {\em Eur. Conf. Comput. Vis.}, 2022.

\bibitem{deng2009imagenet}
Jia Deng, Wei Dong, Richard Socher, Li-Jia Li, Kai Li, and Li Fei-Fei.
\newblock Imagenet: A large-scale hierarchical image database.
\newblock In {\em IEEE Conf. Comput. Vis. Pattern Recog.}, 2009.

\bibitem{dosovitskiy2020image}
Alexey Dosovitskiy, Lucas Beyer, Alexander Kolesnikov, Dirk Weissenborn,
  Xiaohua Zhai, Thomas Unterthiner, Mostafa Dehghani, Matthias Minderer, Georg
  Heigold, Sylvain Gelly, et~al.
\newblock An image is worth 16x16 words: Transformers for image recognition at
  scale.
\newblock In {\em Int. Conf. Learn. Represent.}, 2020.

\bibitem{erichson2022noisymix}
N~Benjamin Erichson, Soon~Hoe Lim, Francisco Utrera, Winnie Xu, Ziang Cao, and
  Michael~W Mahoney.
\newblock Noisymix: Boosting robustness by combining data augmentations,
  stability training, and noise injections.
\newblock {\em arXiv preprint arXiv:2202.01263}, 2022.

\bibitem{everingham2015pascal}
Mark Everingham, SM~Ali Eslami, Luc Van~Gool, Christopher~KI Williams, John
  Winn, and Andrew Zisserman.
\newblock The pascal visual object classes challenge: A retrospective.
\newblock {\em Int. J. Comput. Vis.}, 2015.

\bibitem{pascal-voc-2012}
M. Everingham, L. Van~Gool, C.~K.~I. Williams, J. Winn, and A. Zisserman.
\newblock The {PASCAL} {V}isual {O}bject {C}lasses {C}hallenge 2012 {(VOC2012)}
  {R}esults.
\newblock
  http://www.pascal-network.org/challenges/VOC/voc2012/workshop/index.html.

\bibitem{gatys2016image}
Leon~A Gatys, Alexander~S Ecker, and Matthias Bethge.
\newblock Image style transfer using convolutional neural networks.
\newblock In {\em IEEE Conf. Comput. Vis. Pattern Recog.}, 2016.

\bibitem{geirhos2018}
Robert Geirhos, Patricia Rubisch, Claudio Michaelis, Matthias Bethge, Felix~A
  Wichmann, and Wieland Brendel.
\newblock Imagenet-trained {CNN}s are biased towards texture; increasing shape
  bias improves accuracy and robustness.
\newblock In {\em Int. Conf. Learn. Represent.}, 2019.

\bibitem{gulrajani2020search}
Ishaan Gulrajani and David Lopez-Paz.
\newblock In search of lost domain generalization.
\newblock In {\em Int. Conf. Learn. Represent.}, 2021.

\bibitem{he2015deep}
Kaiming He, Xiangyu Zhang, Shaoqing Ren, and Jian Sun.
\newblock Deep residual learning for image recognition.
\newblock In {\em IEEE Conf. Comput. Vis. Pattern Recog.}, 2015.

\bibitem{hendrycks2021many}
Dan Hendrycks, Steven Basart, Norman Mu, Saurav Kadavath, Frank Wang, Evan
  Dorundo, Rahul Desai, Tyler Zhu, Samyak Parajuli, Mike Guo, Dawn Song, Jacob
  Steinhardt, and Justin Gilmer.
\newblock The many faces of robustness: A critical analysis of
  out-of-distribution generalization.
\newblock In {\em Int. Conf. Comput. Vis.}, 2021.

\bibitem{hendrycks2019robustness}
Dan Hendrycks and Thomas Dietterich.
\newblock Benchmarking neural network robustness to common corruptions and
  perturbations.
\newblock In {\em Int. Conf. Learn. Represent.}, 2019.

\bibitem{hendrycks2019selfsupervised}
Dan Hendrycks, Mantas Mazeika, Saurav Kadavath, and Dawn Song.
\newblock Using self-supervised learning can improve model robustness and
  uncertainty.
\newblock In {\em Adv. Neural Inform. Process. Syst.}, 2019.

\bibitem{hendrycks2019augmix}
Dan Hendrycks, Norman Mu, Ekin~D Cubuk, Barret Zoph, Justin Gilmer, and Balaji
  Lakshminarayanan.
\newblock Augmix: A simple data processing method to improve robustness and
  uncertainty.
\newblock In {\em Int. Conf. Learn. Represent.}, 2020.

\bibitem{hendrycks2021nae}
Dan Hendrycks, Kevin Zhao, Steven Basart, Jacob Steinhardt, and Dawn Song.
\newblock Natural adversarial examples.
\newblock In {\em IEEE Conf. Comput. Vis. Pattern Recog.}, 2021.

\bibitem{howard2019searching_mbv3}
Andrew Howard, Mark Sandler, Grace Chu, Liang-Chieh Chen, Bo Chen, Mingxing
  Tan, Weijun Wang, Yukun Zhu, Ruoming Pang, Vijay Vasudevan, et~al.
\newblock Searching for mobilenetv3.
\newblock In {\em Int. Conf. Comput. Vis.}, 2019.

\bibitem{koh2021wilds}
Pang~Wei Koh, Shiori Sagawa, Henrik Marklund, Sang~Michael Xie, Marvin Zhang,
  Akshay Balsubramani, Weihua Hu, Michihiro Yasunaga, Richard~Lanas Phillips,
  Irena Gao, et~al.
\newblock Wilds: A benchmark of in-the-wild distribution shifts.
\newblock In {\em International Conference on Machine Learning}, 2021.

\bibitem{kortylewski2018empirically}
Adam Kortylewski, Bernhard Egger, Andreas Schneider, Thomas Gerig, Andreas
  Morel-Forster, and Thomas Vetter.
\newblock Empirically analyzing the effect of dataset biases on deep face
  recognition systems.
\newblock In {\em Proceedings of the IEEE Conference on Computer Vision and
  Pattern Recognition Workshops}, 2018.

\bibitem{kortylewski2019analyzing}
Adam Kortylewski, Bernhard Egger, Andreas Schneider, Thomas Gerig, Andreas
  Morel-Forster, and Thomas Vetter.
\newblock Analyzing and reducing the damage of dataset bias to face recognition
  with synthetic data.
\newblock In {\em Proceedings of the IEEE Conference on Computer Vision and
  Pattern Recognition Workshops}, pages 0--0, 2019.

\bibitem{kortylewski2021compositional}
Adam Kortylewski, Qing Liu, Angtian Wang, Yihong Sun, and Alan Yuille.
\newblock Compositional convolutional neural networks: A robust and
  interpretable model for object recognition under occlusion.
\newblock {\em International Journal of Computer Vision}, 2021.

\bibitem{kurakin2016adversarial}
Alexey Kurakin, Ian Goodfellow, and Samy Bengio.
\newblock Adversarial machine learning at scale.
\newblock In {\em Int. Conf. Learn. Represent.}, 2017.

\bibitem{lin2017focal}
Tsung-Yi Lin, Priya Goyal, Ross Girshick, Kaiming He, and Piotr Doll{\'a}r.
\newblock Focal loss for dense object detection.
\newblock In {\em Int. Conf. Comput. Vis.}, 2017.

\bibitem{lin2014microsoft}
Tsung-Yi Lin, Michael Maire, Serge Belongie, James Hays, Pietro Perona, Deva
  Ramanan, Piotr Doll{\'a}r, and C~Lawrence Zitnick.
\newblock Microsoft coco: Common objects in context.
\newblock In {\em Eur. Conf. Comput. Vis.}, 2014.

\bibitem{liu2021swin}
Ze Liu, Yutong Lin, Yue Cao, Han Hu, Yixuan Wei, Zheng Zhang, Stephen Lin, and
  Baining Guo.
\newblock Swin transformer: Hierarchical vision transformer using shifted
  windows.
\newblock In {\em Int. Conf. Comput. Vis.}, 2021.

\bibitem{locatello2020object}
Francesco Locatello, Dirk Weissenborn, Thomas Unterthiner, Aravindh Mahendran,
  Georg Heigold, Jakob Uszkoreit, Alexey Dosovitskiy, and Thomas Kipf.
\newblock Object-centric learning with slot attention.
\newblock In {\em Adv. Neural Inform. Process. Syst.}, 2020.

\bibitem{mahmood2021robustness}
Kaleel Mahmood, Rigel Mahmood, and Marten Van~Dijk.
\newblock On the robustness of vision transformers to adversarial examples.
\newblock In {\em Int. Conf. Comput. Vis.}, 2021.

\bibitem{michaelis2019dragon}
Claudio Michaelis, Benjamin Mitzkus, Robert Geirhos, Evgenia Rusak, Oliver
  Bringmann, Alexander~S. Ecker, Matthias Bethge, and Wieland Brendel.
\newblock Benchmarking robustness in object detection: Autonomous driving when
  winter is coming.
\newblock In {\em Adv. Neural Inform. Process. Syst.}, 2019.

\bibitem{michaelis2019benchmarking}
Claudio Michaelis, Benjamin Mitzkus, Robert Geirhos, Evgenia Rusak, Oliver
  Bringmann, Alexander~S Ecker, Matthias Bethge, and Wieland Brendel.
\newblock Benchmarking robustness in object detection: Autonomous driving when
  winter is coming.
\newblock {\em arXiv preprint arXiv:1907.07484}, 2019.

\bibitem{Mohseni2021PracticalML}
Sina Mohseni, Haotao Wang, Zhiding Yu, Chaowei Xiao, Zhangyang Wang, and Jay
  Yadawa.
\newblock Practical machine learning safety: A survey and primer.
\newblock {\em ArXiv}, 2021.

\bibitem{qiu2016unrealcv}
Weichao Qiu and Alan Yuille.
\newblock Unrealcv: Connecting computer vision to unreal engine.
\newblock In {\em European Conference on Computer Vision}, pages 909--916.
  Springer, 2016.

\bibitem{recht2019imagenet}
Benjamin Recht, Rebecca Roelofs, Ludwig Schmidt, and Vaishaal Shankar.
\newblock Do imagenet classifiers generalize to imagenet?
\newblock In {\em Int. Conf. Machine Learning}, 2019.

\bibitem{ren2015faster}
Shaoqing Ren, Kaiming He, Ross Girshick, and Jian Sun.
\newblock Faster r-cnn: Towards real-time object detection with region proposal
  networks.
\newblock In {\em Adv. Neural Inform. Process. Syst.}, 2015.

\bibitem{rosenfeld2018elephant}
Amir Rosenfeld, Richard Zemel, and John~K Tsotsos.
\newblock The elephant in the room.
\newblock {\em arXiv preprint arXiv:1808.03305}, 2018.

\bibitem{Shao_2021_WACV}
Jie Shao, Xin Wen, Bingchen Zhao, and Xiangyang Xue.
\newblock Temporal context aggregation for video retrieval with contrastive
  learning.
\newblock In {\em IEEE Winter Conf. on Applications of Comput. Vis.}, 2021.

\bibitem{tang2022invariant}
Kaihua Tang, Mingyuan Tao, Jiaxin Qi, Zhenguang Liu, and Hanwang Zhang.
\newblock Invariant feature learning for generalized long-tailed
  classification.
\newblock In {\em Eur. Conf. Comput. Vis.}, 2022.

\bibitem{tremblay2018falling}
Jonathan Tremblay, Thang To, and Stan Birchfield.
\newblock Falling {T}hings: {A} synthetic dataset for {3D} object detection and
  pose estimation.
\newblock In {\em Proceedings of the IEEE Conference on Computer Vision and
  Pattern Recognition Workshops}, 2018.

\bibitem{wang2021nemo}
Angtian Wang, Adam Kortylewski, and Alan Yuille.
\newblock Nemo: Neural mesh models of contrastive features for robust 3d pose
  estimation.
\newblock In {\em Int. Conf. Learn. Represent.}, 2021.

\bibitem{wang2020robust}
Angtian Wang, Yihong Sun, Adam Kortylewski, and Alan~L Yuille.
\newblock Robust object detection under occlusion with context-aware
  compositionalnets.
\newblock In {\em IEEE Conf. Comput. Vis. Pattern Recog.}, 2020.

\bibitem{wang2021augmax}
Haotao Wang, Chaowei Xiao, Jean Kossaifi, Zhiding Yu, Anima Anandkumar, and
  Zhangyang Wang.
\newblock Augmax: Adversarial composition of random augmentations for robust
  training.
\newblock In {\em NeurIPS}, 2021.

\bibitem{wen2022selfsupervised}
Xin Wen, Bingchen Zhao, Anlin Zheng, Xiangyu Zhang, and Xiaojuan Qi.
\newblock Self-supervised visual representation learning with semantic
  grouping.
\newblock {\em arxiv: 2205.15288}, 2022.

\bibitem{wong2020fast}
Eric Wong, Leslie Rice, and J~Zico Kolter.
\newblock Fast is better than free: Revisiting adversarial training.
\newblock In {\em Int. Conf. Learn. Represent.}, 2020.

\bibitem{xiang2014beyond}
Yu Xiang, Roozbeh Mottaghi, and Silvio Savarese.
\newblock Beyond pascal: A benchmark for 3d object detection in the wild.
\newblock In {\em IEEE Winter Conf. on Applications of Comput. Vis.}, 2014.

\bibitem{xiang_wacv14}
Yu Xiang, Roozbeh Mottaghi, and Silvio Savarese.
\newblock Beyond pascal: A benchmark for 3d object detection in the wild.
\newblock In {\em IEEE Winter Conf. on Applications of Comput. Vis.}, 2014.

\bibitem{xiang2017posecnn}
Yu Xiang, Tanner Schmidt, Venkatraman Narayanan, and Dieter Fox.
\newblock Posecnn: A convolutional neural network for 6d object pose estimation
  in cluttered scenes.
\newblock In {\em Robotics: Science and Systems (RSS)}, 2018.

\bibitem{xiao2020tdmpnet}
Mingqing Xiao, Adam Kortylewski, Ruihai Wu, Siyuan Qiao, Wei Shen, and Alan
  Yuille.
\newblock Tdmpnet: Prototype network with recurrent top-down modulation for
  robust object classification under partial occlusion.
\newblock In {\em European Conference on Computer Vision}, pages 447--463.
  Springer, 2020.

\bibitem{xie2019feature}
Cihang Xie, Yuxin Wu, Laurens van~der Maaten, Alan~L Yuille, and Kaiming He.
\newblock Feature denoising for improving adversarial robustness.
\newblock In {\em IEEE Conf. Comput. Vis. Pattern Recog.}, 2019.

\bibitem{ye2021ood}
Nanyang Ye, Kaican Li, Lanqing Hong, Haoyue Bai, Yiting Chen, Fengwei Zhou, and
  Zhenguo Li.
\newblock Ood-bench: Benchmarking and understanding out-of-distribution
  generalization datasets and algorithms.
\newblock {\em arXiv preprint arXiv:2106.03721}, 2021.

\bibitem{yun2019cutmix}
Sangdoo Yun, Dongyoon Han, Seong~Joon Oh, Sanghyuk Chun, Junsuk Choe, and
  Youngjoon Yoo.
\newblock Cutmix: Regularization strategy to train strong classifiers with
  localizable features.
\newblock In {\em Int. Conf. Comput. Vis.}, 2019.

\bibitem{zhao2020distilling}
Bingchen Zhao and Xin Wen.
\newblock Distilling visual priors from self-supervised learning.
\newblock In {\em Eur. Conf. Comput. Vis.}, 2020.

\bibitem{zhou2018starmap}
Xingyi Zhou, Arjun Karpur, Linjie Luo, and Qixing Huang.
\newblock Starmap for category-agnostic keypoint and viewpoint estimation.
\newblock In {\em Eur. Conf. Comput. Vis.}, 2018.

\bibitem{zhu2021improving}
Rui Zhu, Bingchen Zhao, Jingen Liu, Zhenglong Sun, and Chang~Wen Chen.
\newblock Improving contrastive learning by visualizing feature transformation.
\newblock In {\em Int. Conf. Comput. Vis.}, 2021.

\end{thebibliography}


\begin{thebibliography}{10}\itemsep=-1pt

\bibitem{everingham2015pascal}
Mark Everingham, SM~Ali Eslami, Luc Van~Gool, Christopher~KI Williams, John
  Winn, and Andrew Zisserman.
\newblock The pascal visual object classes challenge: A retrospective.
\newblock {\em Int. J. Comput. Vis.}, 2015.

\bibitem{geirhos2018}
Robert Geirhos, Patricia Rubisch, Claudio Michaelis, Matthias Bethge, Felix~A
  Wichmann, and Wieland Brendel.
\newblock Imagenet-trained {CNN}s are biased towards texture; increasing shape
  bias improves accuracy and robustness.
\newblock In {\em Int. Conf. Learn. Represent.}, 2019.

\bibitem{he2015deep}
Kaiming He, Xiangyu Zhang, Shaoqing Ren, and Jian Sun.
\newblock Deep residual learning for image recognition.
\newblock In {\em IEEE Conf. Comput. Vis. Pattern Recog.}, 2015.

\bibitem{hendrycks2019augmix}
Dan Hendrycks, Norman Mu, Ekin~D Cubuk, Barret Zoph, Justin Gilmer, and Balaji
  Lakshminarayanan.
\newblock Augmix: A simple data processing method to improve robustness and
  uncertainty.
\newblock In {\em Int. Conf. Learn. Represent.}, 2020.

\bibitem{howard2019searching_mbv3}
Andrew Howard, Mark Sandler, Grace Chu, Liang-Chieh Chen, Bo Chen, Mingxing
  Tan, Weijun Wang, Yukun Zhu, Ruoming Pang, Vijay Vasudevan, et~al.
\newblock Searching for mobilenetv3.
\newblock In {\em Int. Conf. Comput. Vis.}, 2019.

\bibitem{lin2017focal}
Tsung-Yi Lin, Priya Goyal, Ross Girshick, Kaiming He, and Piotr Doll{\'a}r.
\newblock Focal loss for dense object detection.
\newblock In {\em Int. Conf. Comput. Vis.}, 2017.

\bibitem{liu2021swin}
Ze Liu, Yutong Lin, Yue Cao, Han Hu, Yixuan Wei, Zheng Zhang, Stephen Lin, and
  Baining Guo.
\newblock Swin transformer: Hierarchical vision transformer using shifted
  windows.
\newblock In {\em Int. Conf. Comput. Vis.}, 2021.

\bibitem{ren2015faster}
Shaoqing Ren, Kaiming He, Ross Girshick, and Jian Sun.
\newblock Faster r-cnn: Towards real-time object detection with region proposal
  networks.
\newblock In {\em Adv. Neural Inform. Process. Syst.}, 2015.

\bibitem{wang2021nemo}
Angtian Wang, Adam Kortylewski, and Alan Yuille.
\newblock Nemo: Neural mesh models of contrastive features for robust 3d pose
  estimation.
\newblock In {\em Int. Conf. Learn. Represent.}, 2021.

\bibitem{wong2020fast}
Eric Wong, Leslie Rice, and J~Zico Kolter.
\newblock Fast is better than free: Revisiting adversarial training.
\newblock In {\em Int. Conf. Learn. Represent.}, 2020.

\bibitem{zhou2018starmap}
Xingyi Zhou, Arjun Karpur, Linjie Luo, and Qixing Huang.
\newblock Starmap for category-agnostic keypoint and viewpoint estimation.
\newblock In {\em Eur. Conf. Comput. Vis.}, 2018.

\end{thebibliography}
}

%\clearpage
\end{document}

%%%%%%%%% TITLE - PLEASE UPDATE
%\title{Appendix: \OURS: A Benchmark for Robustness to Individual Nuisances \\in Real-World Out-of-Distribution Shifts}
%\title{\OURS: A Benchmark for Robustness to Semantic Out-of-Distribution Shifts in the Real World}

%\author{Bingchen Zhao$^{1,5}$,\hspace{0.3em} Shaozuo Yu$^{1,5}$,\hspace{0.3em} Wufei Ma$^3$,\hspace{0.3em} Mingxin Yu$^4$,\hspace{0.3em} Shenxiao Mei$^2$,\\ Angtian Wang$^2$,\hspace{0.3em} Ju He$^2$,\hspace{0.3em} Alan Yuille$^2$,\hspace{0.3em} Adam Kortylewski$^2$\\
%$^1$Tongji University\hspace{0.2em} $^2$ Johns Hopkins University\hspace{0.2em} $^3$ Purdue University\hspace{0.2em}  $^4$ Peking University\hspace{0.2em} $^5$ LunarAI\\
%{\tt\small \{zhaobc.gm, yu.shaozuo\}@gmail.com, \{ayuille1, akortyl1\}@jhu.edu}
%}
%\maketitle

\appendix

\section{Implementation Details}
In this section, we introduce the implementations details of the models and techniques for improving the robustness in the experiments conducted in Sec.~4. 

\subsection{Image Classification}
For the experiments of image classification on \OURS datasets, we tested three network architectures, namely, MobileNetV3-Large~\cite{howard2019searching_mbv3}, ResNet-50~\cite{he2015deep}, and Swin-T~\cite{liu2021swin}.
We train all the three models with the same hyper-parameter to make a fair comparison.
The Batchsize is set to 64 with a step decayed learning rate initialized with 1e-4 and then multiplied by 15,30,45 epochs, we train the network for a total of 100 epochs on the training set.
The resolution of the input images are 224 by 224 which is also a defaulted value for training networks~\cite{he2015deep}.

We compared the effectiveness of different data augmentation techniques, namely, style transfer~\cite{geirhos2018}, AugMix~\cite{hendrycks2019augmix}, and adversarial training~\cite{wong2020fast}.
For all the experiments using style transfer~\cite{geirhos2018}, we use the code from the original authors~\footnote{https://github.com/rgeirhos/Stylized-ImageNet} to create the style augmented images for training.
For experiments with AugMix~\cite{hendrycks2019augmix}, we adopted a PyTorch-based implementation~\footnote{https://github.com/psh150204/AugMix}. 
For adversarial training, we adopted the implementation from the official source.~\footnote{https://github.com/locuslab/fast\_adversarial}

\subsection{Object Detection}
We mainly used two frameworks for the task of object detection, namely Faster-RCNN~\cite{ren2015faster} and RetinaNet~\cite{lin2017focal}.
Similarly, we keep all the hyper-parameter the same except for the ones we wish to study.
The experiments are mainly conducted using the detectron2 codebase~\footnote{https://github.com/facebookresearch/detectron2}.
For strong data augmentation techniques that can be used to improve the robustness of vision models, AugMix~\cite{hendrycks2019augmix} is relatively harder to implement than the other on object detection because of the image mixing step, so we only evaluated the performance of style transfer and adversarial training.
The style transfer uses the same images generate for image classification, and we followed the same procedure to do the adversarial training for object detection.

We train all the object detection models with 18000 iterations with a initial learning rate of 0.02 and a batchsize of 16, the learning rate is then multiplied by 0.1 at 12000 and 16000 iterations.
We adopted the multi-scale training technique to improve the baseline performance, each input images will be resized to have a short edge of $[480, 512, 544, 576, 608, 640, 672, 704, 736, 768, 800]$, and when testing, the test input image will be resized to have a short edge of 800.
For experiments with Swin-T as the backbone network in the detection framework, we adopted the implementations from the authors of the swin-transformer~\footnote{https://github.com/SwinTransformer/Swin-Transformer-Object-Detection}.

\subsection{3D pose estimation}
For 3D pose estimation, we evaluated two types of models, Res50-Specific~\cite{zhou2018starmap} and NeMo~\cite{wang2021nemo}.
We adopted the implementation from the original authors~\footnote{https://github.com/shubhtuls/ViewpointsAndKeypoints}\footnote{https://github.com/Angtian/NeMo}.
When training the pose estimation models, we use a batchsize of 108 and a learning rate of 1e-3.
For the pose estimation model for each category, we train the model for 800 epochs.

\section{Detailed statistics}

In~\cref{tab:number_images_category_bias}, we provide the statistics of our dataset. Note that for chair, diningtable, and sofa, it is difficult to find images with the weather nuisance, so the number of images for these categories with weather nuisances is 0.

% the number of images is not balanced in terms of nuisances...
% so maybe not include this table.
%Detailed statistics will be put in our supplementary materials.
\begin{table*}[ht]
\centering
\caption{\label{tab:number_images_category_bias} Number of images in each categories with individual nuisances that we defined.}%\TODO{Caption, collect the numbers from the dataset}}
\begin{tabular}{lrrrrrr}
\toprule
\#img       & Shape  & Pose    & Texture   & Context & Weather & Total \\
\midrule
aeroplane   &  27         &   40         &  66             &  79          &   108          &   320    \\
bus         &  83         &   18         &  82             &  4            &   30          &    217   \\
car         &  159         &   24         &  40             &  20           &   83          &   326    \\
train       &  34        &   42         &  130            &  70           &   66          &    342   \\
boat        &  30         &   82         &  29             &  30           &   76          &   247    \\
bicycle     &  64         &   70         &  28             &  78           &   113          &  353      \\
motorbike   &  89         &   108         &  76             &  27           &   97           & 397      \\
chair       &  40         &   40         &  42             &  17           &   0          &  139     \\
diningtable &  22         &   65         &  18              &  59           &   0          & 164      \\
sofa        &  15         &   28         &  24             &  60          &   0           &  127     \\
\midrule
Total       &  563         &   517         &  535      &  444    &   573          &   2632    \\
\bottomrule
\end{tabular}
\end{table*}

\section{Example Images from \OURS}
We show some example images next page. We will release the full dataset.%~\footnote{https://anonymous.4open.science/r/ROBIN-supplementary-BC7F/README.md}. 

\begin{figure*}[htbp]
\includegraphics[width=\linewidth]{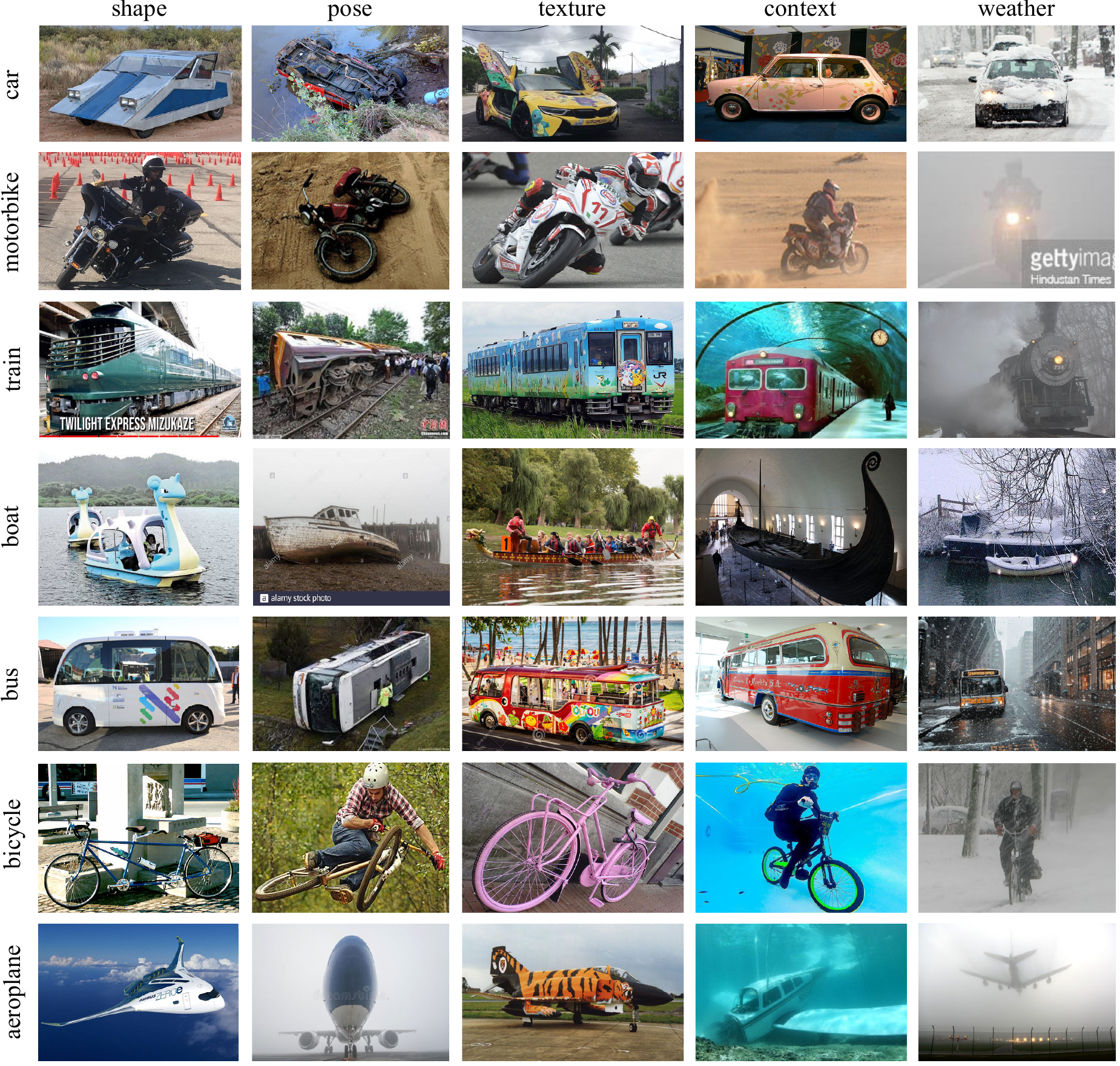}
\caption{\label{fig:supp_images} More example images from \OURS dataset, we will release the full dataset.}
\end{figure*}

\section{Images filtered from the original PASCAL3D+ dataset}
This section shows example images that we filtered out from the original PASCAL3D+ dataset~\cite{everingham2015pascal} in order to make the \OURS test set really OOD.
The images are removed because they are too similar to the images in the \OURS test set.

In our anonymous repository, we provide all the images that we removed from the original PASCAL3D+ dataset.

\begin{figure}[htbp]
\includegraphics[width=\linewidth]{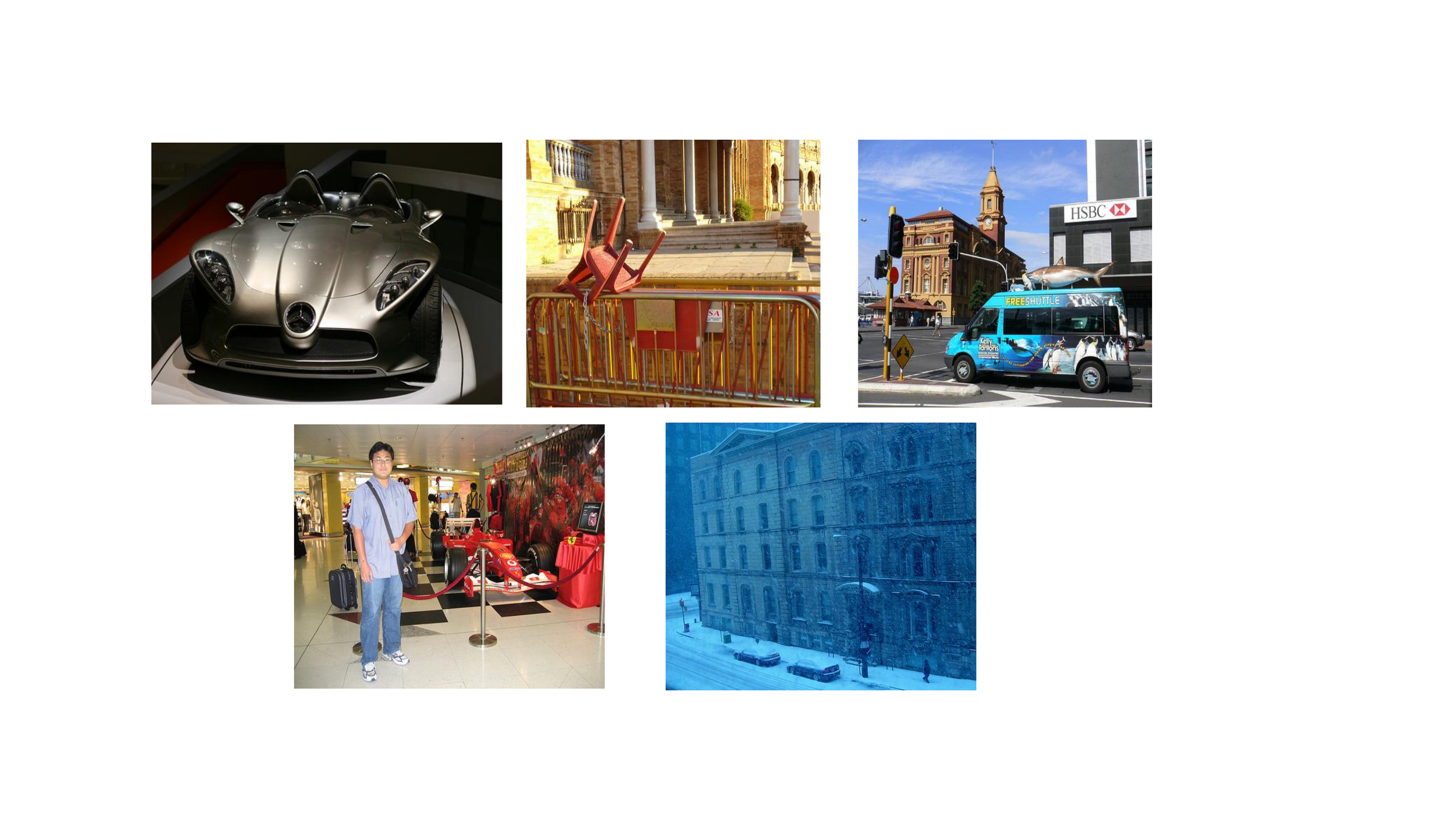}
\caption{\label{fig:filter_pascal} Example images that are filtered out from the original PASCAL3D+ dataset. These images has nuisances that are similar to the ones we collected in the \OURS dataset, so they are removed from the training set. We attached all the filtered images with the supplementary.}
\end{figure}

\section{Example images with multiple nuisances}
\begin{figure}[htbp]
\includegraphics[width=\linewidth]{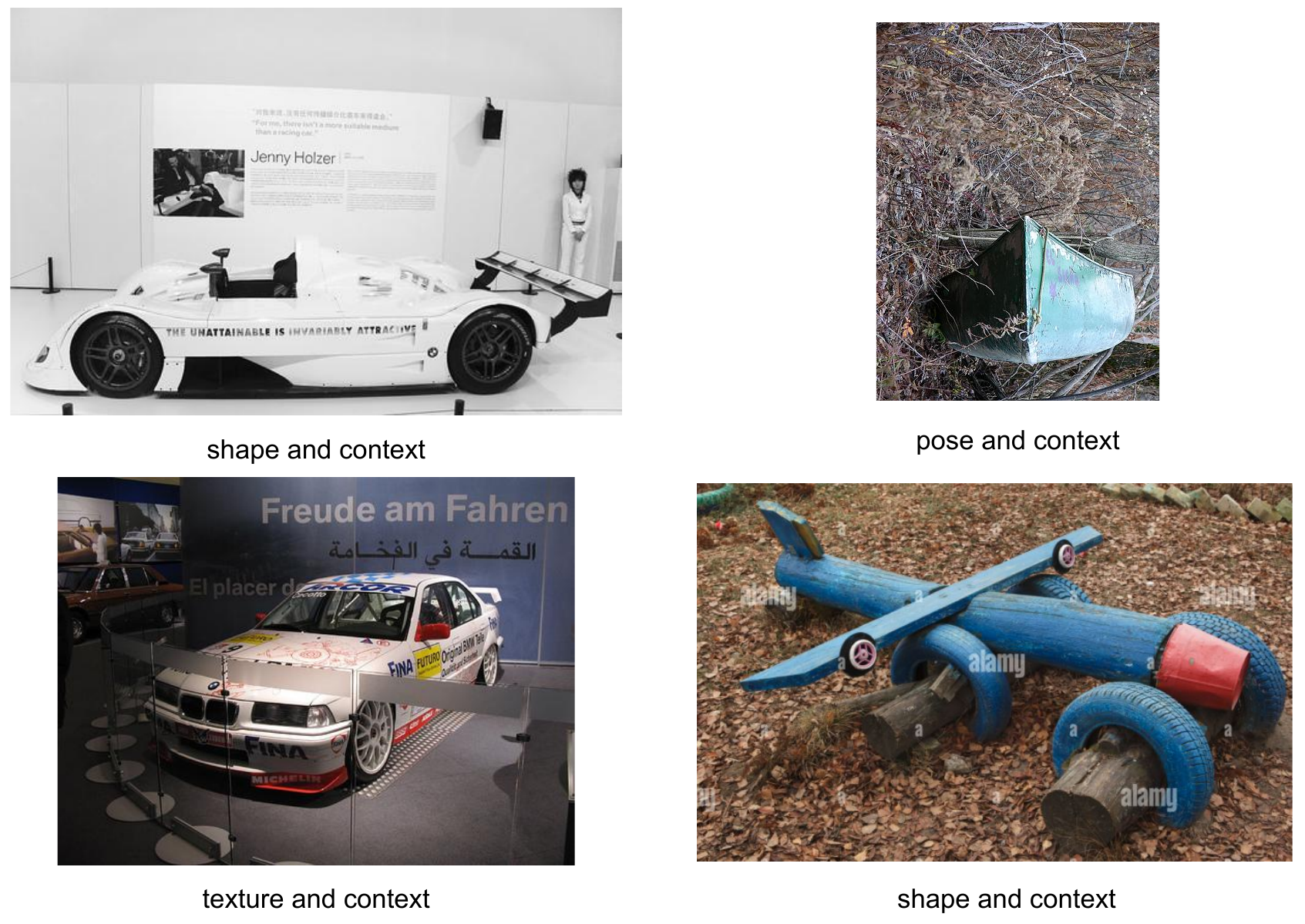}
\caption{\label{fig:multiple_nuisance} Example images with multiple nuisance. From our internet search, we also collected many images with multiple nuisance factors, these images are later removed to ensure that we are testing with only one controllable nuisances.}
\end{figure}

We also removed the images that have multiple nuisances from our internet search, we give examples of multiple nuisances in~\cref{fig:multiple_nuisance}.

\section{The user interface of our annotation tools}
Here we also provide the user interface of our used annotation tools for bounding boxes annotation and 3D pose annotations.
The annotation tools are taken and slightly modified from a GitHub project~\footnote{https://github.com/jsbroks/coco-annotator} and the original PASCAL3D+ dataset~\footnote{https://cvgl.stanford.edu/projects/pascal3d.html}. Identifying informations have been removed from the screenshots.

\begin{figure}[htbp]
\includegraphics[width=\linewidth]{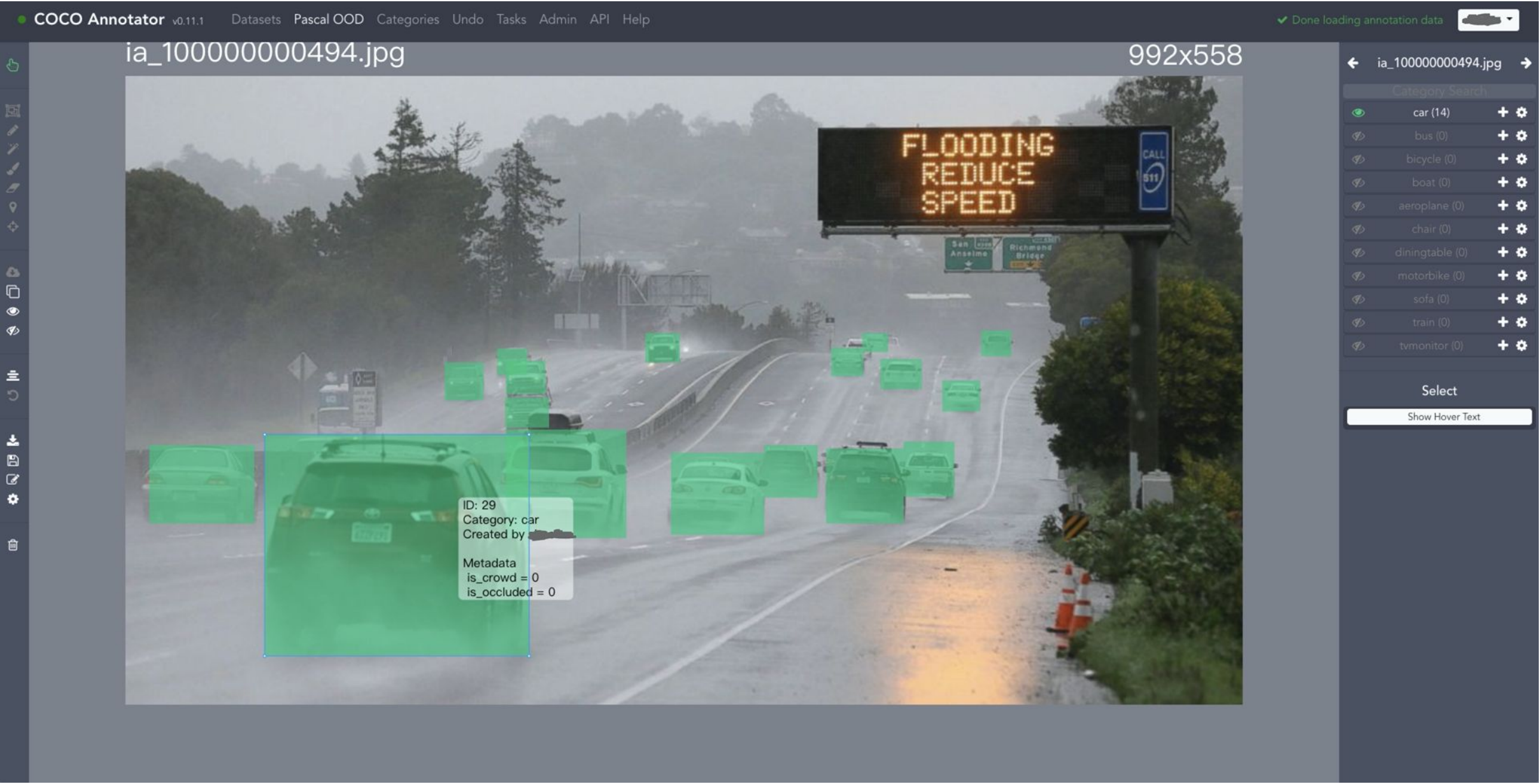}
\caption{\label{fig:det_ui} The user interface of the detection annotation tool.}
\end{figure}

\begin{figure}[htbp]
\includegraphics[width=\linewidth]{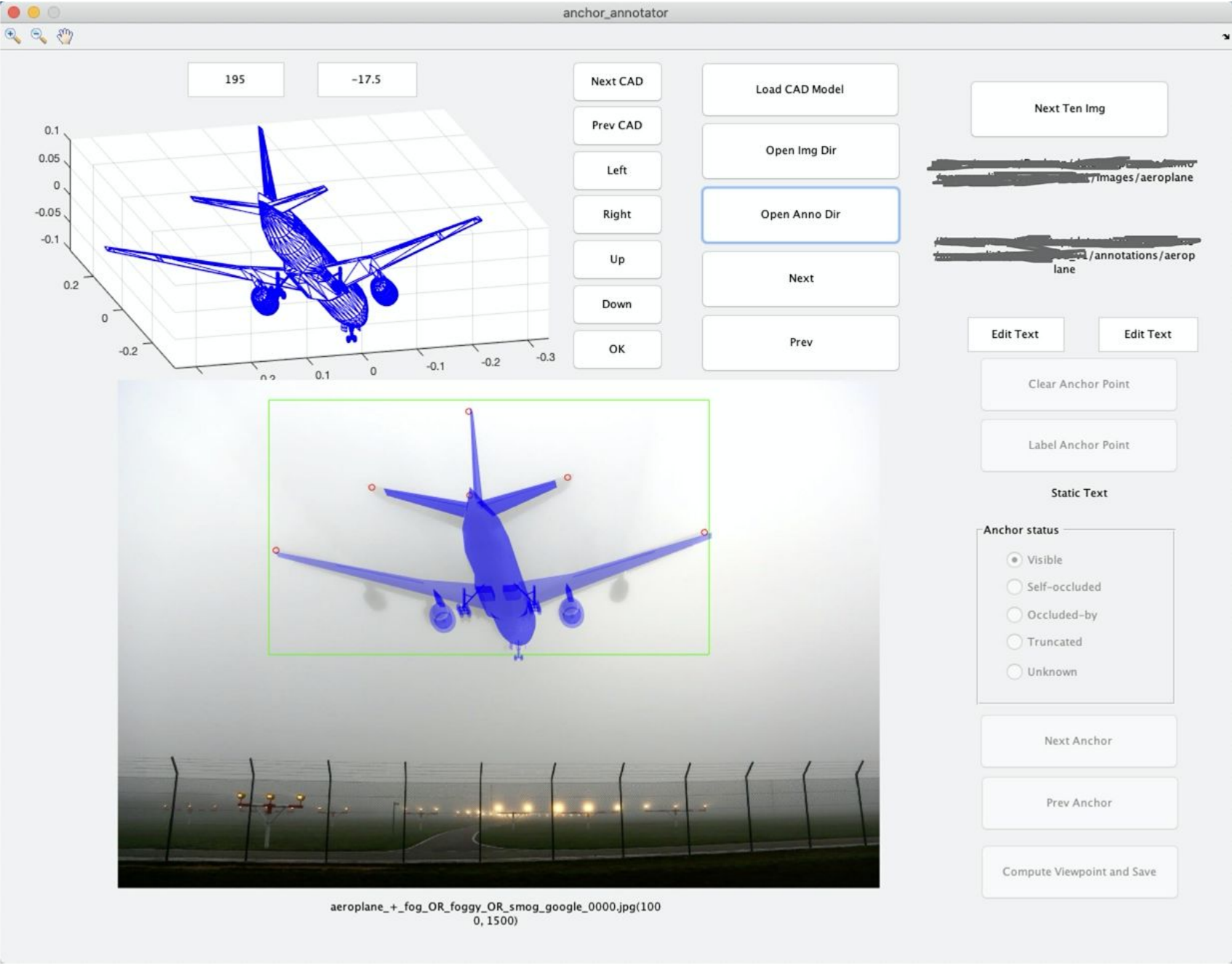}
\caption{\label{fig:det_ui} The user interface of the 3D pose annotation tool.}
\end{figure}

%%%%%%%%% REFERENCES
{
\small
\bibliographystyle{ieee_fullname}
\bibliography{eccv_egbib}
}